\documentclass[preprint,12pt]{elsarticle}

\usepackage{xcolor}
\usepackage{lscape}
\usepackage{amsthm}
\usepackage[cmex10]{amsmath}
\usepackage{amsfonts}
\usepackage{amssymb}
\usepackage{siunitx}
\usepackage{textcomp}
\usepackage{booktabs}
\usepackage{graphicx}

\usepackage{subfigure}

\usepackage{algorithm}
\usepackage{algorithmic}
\usepackage{microtype}
\usepackage{longtable}
\usepackage{placeins}
\usepackage{caption}
\usepackage{setspace}
\usepackage{tabularx}

\usepackage{hyperref}

\journal{Sensors and Actuators A: Physical}

\newtheorem{highlight}{Highlight}

\begin{document}

\onehalfspacing

\begin{frontmatter}

\title{Design and Characteristics of a Thin-Film ThermoMesh for the Efficient Embedded Sensing of a Spatio-Temporally Sparse Heat Source}

\author{Sajjad Boorghan Farahan}
\address{Department of Mechanical Engineering\\ State University of New York at Binghamton, Binghamton, NY}
\author{Ahmed Alajlouni}
\address{Department of Mechanical Engineering\\ State University of New York at Binghamton, Binghamton, NY}
\author{Jingzhou Zhao*}
\address{Department of Mechanical Engineering\\ State University of New York at Binghamton, Binghamton, NY}
\cortext[Jingzhou Zhao]{Corresponding author}
\ead{jingzhou.zhao@binghamton.edu}


\begin{highlight}
A new thin-film ThermoMesh that simultaneously senses and compresses a spatio-temporally sparse temperature signal passively using phonon and electron transport.
\end{highlight}

\begin{highlight}
Performance characteristics (range, efficiency, sensitivity, and accuracy) are defined and derived based on an experimentally validated model to arrive at practical designs for both contact and non-contact conduction-based thermal imaging systems.
\end{highlight}

\begin{highlight}
The new sensor promises orders of magnitude improvement in energy efficiency for the measurement of extremely rare thermal events that are sparse in both space and time. 
\end{highlight}

\begin{abstract}

This work presents ThermoMesh, a passive thin-film thermoelectric mesh sensor designed to detect and characterize spatio-temporally sparse heat sources through conduction-based thermal imaging. The device integrates thermoelectric junctions with linear or nonlinear interlayer resistive elements to perform simultaneous sensing and in-sensor compression. We focus on the single-event (1-sparse) operation and define four performance metrics: range, efficiency, sensitivity, and accuracy. Numerical modeling shows that a linear resistive interlayer flattens the sensitivity distribution and improves minimum sensitivity by approximately tenfold for a $16\times16$ mesh. Nonlinear temperature-dependent interlayers further enhance minimum sensitivity at scale: a ceramic negative-temperature-coefficient (NTC) layer over 973--1273~K yields a $\sim14{,}500\times$ higher minimum sensitivity than the linear design at a $200\times200$ mesh, while a VO$_2$ interlayer modeled across its metal--insulator transition (MIT) over 298--373~K yields a $\sim24\times$ improvement. Using synthetic 1-sparse datasets with white boundary-channel noise at a signal-to-noise ratio of 40~dB, the VO$_2$ case achieved $98\%$ localization accuracy, a mean absolute temperature error of $0.23$~K, and a noise-equivalent temperature (NET) of $0.07$~K. For the ceramic-NTC case no localization errors were observed under the tested conditions, with a mean absolute temperature error of $1.83$~K and a NET of $1.49$~K. These results indicate that ThermoMesh could enable energy-efficient embedded thermal sensing in scenarios where conventional infrared imaging is limited, such as molten-droplet detection or hot-spot monitoring in harsh environments.

\end{abstract}


\end{frontmatter}


\section{Introduction and Background}\label{sec:introduction}

Many context-specific imaging tasks (e.g., localization, anomaly detection, pattern recognition, etc.) benefit from signal compression in the physical, analog, or digital domain~\cite{pope2013light,liang2024physical,wan2023sensor,zhao2024sensor} to achieve significant performance improvements at the system level. This has been successfully demonstrated for magnetic~\cite{lustig2007sparse}, tactile~\cite{li2021inorganic,akhtar2015efficient,luo2012compressive,cao2025programmable,li2026massively}, acoustic~\cite{lin2025blind}, electromagnetic~\cite{ender2010compressive,anitori2011detection}, and optical~\cite{huang2025sensor} imaging, to name a few. However, very few works~\cite{zhao2019thermomesh,silva2026thermal} have been proposed or reported to date that perform in-sensor compression based on phonon and electron transport for applications that require conduction-based thermal imaging, presenting a significant untapped potential.

Conduction-based thermal imaging systems typically rely on the minute diffusion length or thermal mass of a thin-film to achieve the desired temporal resolution in both contact ~\cite{liu2024recent,hidalgo2018low,huang2013flexible,katerinopoulou2019large,urban1990high,bucher2022printed,liu2024flexible,nakajima2020ultrathin,webb2013ultrathin,daus2022fast} and non-contact~\cite{yadav2022advancements} use cases. Employing the emerging physical in-sensor computing~\cite{wan2023sensor} and compression~\cite{huang2025sensor} concepts promises new designs with enlarged sensing area, reduced power consumption, and simplified readout circuitry for many high-stakes applications that require the timely measurement of a rare or spatio-temporally sparse temperature signal, such as the detection of Lithium-ion battery thermal runaway~\cite{luo2025review,huang2021review}, monitoring of laser- and particle-based thermal processes~\cite{sala2024ai,johnson2024monitoring}, detection of spatio-temporally resolved hot-spot in microelectronics ~\cite{de2000nucleation,liu2024thermoelectric}, understanding of laminar-turbulent boundary layer transition ~\cite{yang2023high,shadloo2017laminar}, bolometer-based radar~\cite{simoens2014terahertz}, detection of solar-flare, and artificial thermal nociceptors, etc. Using sparsity explicitly as a design prior is well-motivated because sparse representations are widely studied as an efficient strategy for encoding structured signals~\cite{olshausen2004sparse}.

To this end, we propose an analog--digital thin-film thermal mesh sensor (i.e., ThermoMesh) that utilizes thermoelectric and thermally responsive nonlinear resistive elements (e.g., thermal switches or thermistors) to perform simultaneous sensing and computing passively in the physical domain, with read-only resistive memories programmed into the structure and material properties of the device at the manufacturing stage. Using an experimentally validated model \cite{zhao2019thermomesh} adapted to this new design, we progressively arrive at acceptable performance characteristics for practically measuring the location and magnitude of spatio-temporally sparse temperature signals (e.g., generated by a small and fast-moving molten metal droplet both before and after impact \cite{escure2003experimental}). 

Section~\ref{sec:Design and Performance} defines the ThermoMesh architecture and performance metrics, and reports methods and results for linear and nonlinear interlayer designs. Section~\ref{sec:Discussion} provides an in-depth discussion of design implications, limitations, and potential for broader integration. Finally, Section~\ref{sec:Conclusion} concludes the paper and outlines future directions, including fabrication, experimental validation, and potential applications.

\section{System Design and Performance Characteristics}\label{sec:Design and Performance}

In this section, we present the main methods and results of this work, divided into four parts. First, we define the discrete ThermoMesh measurement model—showing how it \emph{simultaneously senses and compresses} spatio-temporally sparse inputs—by defining the sensitivity matrix $\mathbf{A}$ that maps the interior temperature vector $\mathbf{T}$ to the perimeter boundary-voltage vector $\mathbf{V}$, and we summarize the performance metrics used throughout (i.e., range, efficiency, sensitivity, and accuracy). Second, we present an analog--digital thin-film sensing architecture for 2D heat-source localization and temperature estimation under a single-pixel operating regime, which motivates the ideal-switch interpretation and the interlayer variants studied in this paper. Third, we assemble the \emph{linear} sensing matrix and analyze its properties, quantifying per-junction and minimum sensitivities and studying how a temperature-independent interlayer resistance reshapes the sensitivity distribution. Finally, we extend the model by incorporating a \emph{temperature-dependent} interlayer (e.g., negative-temperature-coefficient (NTC) materials), for which the mapping becomes state dependent; we evaluate its impact on minimum sensitivity and super-linear behavior across operating temperatures.

\subsection{Sensor designs and performance metrics}
\label{sec:design-metrics}

\begin{figure}[!t]
  \centering
  \includegraphics[width=\linewidth,height=\textheight,keepaspectratio]{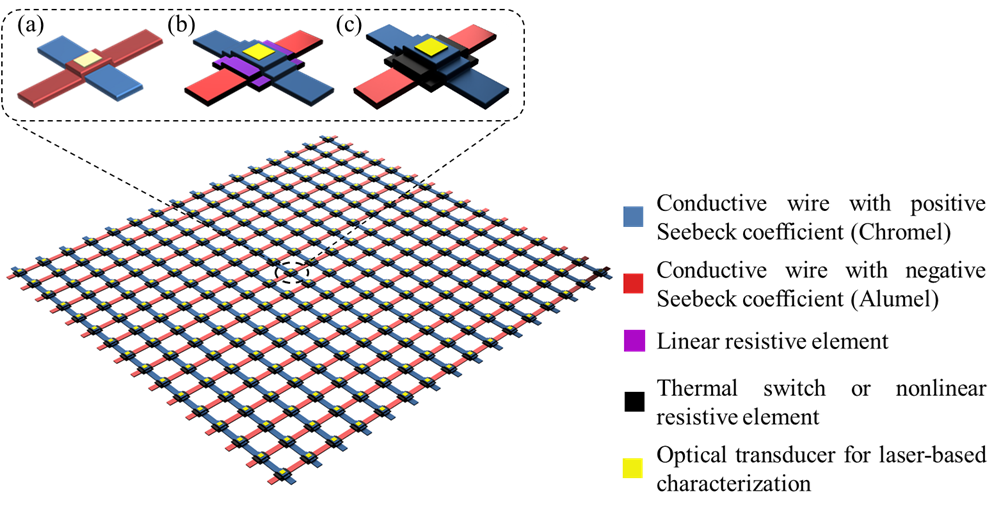}%
  \caption{\textbf{ThermoMesh physical architecture and interlayer variants.}
  A mesh of orthogonal \textbf{Chromel} and \textbf{Alumel} lines forms an array of thermocouple junctions. The inset highlights three junction configurations considered in this work: (a) a direct metal–metal crossing without an added interlayer, (b) insertion of a temperature-independent resistive interlayer yielding linear behavior, and (c) incorporation of a temperature-dependent interlayer that produces nonlinear or switch-like behavior.}
  \label{fig:tm-variants}
\end{figure}

The ThermoMesh sensor is composed of thermocouple wires assembled (or welded, printed, deposited, etc.) in a wire mesh configuration. Their crossings form thermocouple junctions, which are the physical temperature-sensing sites of the device and are assumed to be in thermal equilibrium with the measurand. When the sensor is viewed as an imaging array, each sensing junction is treated as one pixel. By contrast, in the electrical network model, a node denotes an electrical unknown used in the Kirchhoff current law (KCL) formulation. Temperature differences between adjacent junctions generate thermoelectric electromotive force (EMF) in the thermocouple wires (Chromel and Alumel for K-type thermocouples) due to the Seebeck effect. The wire segments connecting adjacent nodes can therefore be considered as voltage sources with internal resistance, whose EMF is determined by the temperature difference and the Seebeck coefficient of the wires between adjacent nodes. Fig.~\ref{fig:tm-variants} summarizes the three interlayer variants analyzed in this work. In the figure, the top-layer patch is a gold pad used only as an optical transducer for laser-based characterization.

We work with a discrete temperature signal defined on the junction grid. Let the array have $M$ rows and $N$ columns of junctions, and let $n=1,2,\ldots$ index discrete sampling times $t_n$. The temperature at junction $(i,j)$ and frame $n$ is denoted
\begin{equation}
  T_{i,j}[n], 
  \qquad i=1,\ldots,M,\;\; j=1,\ldots,N,
\end{equation}
and we collect all $MN$ junction temperatures at frame $n$ into a column vector
\begin{equation}
  \mathbf{T}[n]
  \;=\;
  \big[\,T_{1,1}[n],\ldots,T_{1,N}[n],T_{2,1}[n],\ldots,T_{M,N}[n]\,\big]^{\mathsf T}
  \in\mathbb{R}^{MN}.
\end{equation}
This $\mathbf{T}[n]$ is the discrete temperature signal that the sensor is designed to measure at frame $n$. A reconstruction method produces an estimate $\hat{\mathbf{T}}[n]$ from the measured boundary voltages; later sections evaluate performance by comparing $\hat{\mathbf{T}}[n]$ with $\mathbf{T}[n]$.

At each frame, the exterior boundary voltages are collected into a vector
\begin{equation}
  \mathbf{V}[n]
  \;=\;
  \big[V_1[n],\ldots,V_{2M+2N}[n]\big]^{\mathsf T}
  \in\mathbb{R}^{2M+2N},
\end{equation}
where the index runs over all boundary channels on the perimeter of the mesh. For a given ThermoMesh design, the relationship between $\mathbf{V}[n]$ and $\mathbf{T}[n]$ is obtained from a network model. In the linear (temperature-independent) case this is a matrix--vector product
\begin{equation}
  \mathbf{V}[n] \;=\; \mathbf{A}\,\mathbf{T}[n],
  \qquad
  \mathbf{A}\in\mathbb{R}^{(2M+2N)\times(MN)},
  \label{eq:boundary_map}
\end{equation}
where $\mathbf{A}$ is the \emph{sensitivity matrix} that maps the interior temperature vector to the measured boundary-voltage vector. With a temperature-dependent interlayer, the coefficients of $\mathbf{A}$ depend on the operating temperatures, and the measurement model becomes
\begin{equation}
  \mathbf{V}[n] \;=\; \mathbf{A}\!\big(\mathbf{T}[n]\big)\,\mathbf{T}[n].
  \label{eq:nonlinear-forward}
\end{equation}

These relations hold independently at each time stamp $t_n$ as long as charge transport is much faster than heat transport. In the parallel-plate limit, the RC time constant of each junction is approximately equal to $\rho \varepsilon_0 \varepsilon_r$, where $\rho$ is the interlayer resistivity, $\varepsilon_r$ is the interlayer relative permittivity, and $\varepsilon_0$ is the permittivity of free space. It is therefore independent of pixel dimensions and interlayer geometry and is set primarily by material properties. For the interlayer materials considered in this work, it is estimated to be smaller than $1~\mu\mathrm{s}$, so its effect on response time is neglected in the present model; by contrast, materials with much larger resistivity and permittivity (e.g., pyroelectric materials) could yield a larger RC time constant and correspondingly slower response.

The matrix $\mathbf{A}$ is computed by applying KCL to the network and using the wire and interlayer material relations. For clarity, we consider the 3D stack indexed by $(i,j,k)$ for row, column, and layer. At each interior node $(i,j,k)$, KCL gives

\begin{equation}
\begin{alignedat}{2}
& G_{i,j,k;i-1,j,k}(U_{i,j,k}-U_{i-1,j,k})
&\;+\;& G_{i,j,k;i+1,j,k}(U_{i,j,k}-U_{i+1,j,k}) \\
& + G_{i,j,k;i,j-1,k}(U_{i,j,k}-U_{i,j-1,k})
&\;+\;& G_{i,j,k;i,j+1,k}(U_{i,j,k}-U_{i,j+1,k}) \\
& + G_{i,j,k;i,j,k-1}(U_{i,j,k}-U_{i,j,k-1})
&\;+\;& G_{i,j,k;i,j,k+1}(U_{i,j,k}-U_{i,j,k+1}) \\[6pt]
=\;& -S_{i,j,k;i-1,j,k}(T_{i,j}-T_{i-1,j})
&\;-\;& S_{i,j,k;i+1,j,k}(T_{i,j}-T_{i+1,j}) \\
& - S_{i,j,k;i,j-1,k}(T_{i,j}-T_{i,j-1})
&\;-\;& S_{i,j,k;i,j+1,k}(T_{i,j}-T_{i,j+1})
\end{alignedat}
\label{eq:KCL-3D}
\end{equation}

where $G_{\cdot}$ are inter-node conductances, $S_{\cdot}$ are the effective Seebeck couplings, $U_{i,j,k}$ is the electric potential of node $(i,j,k)$, and $T_{i,j,k}$ is the corresponding junction temperature relative to the cold junction. Peltier and Thomson effects are neglected assuming small currents ($<$1~$\mu$A). In matrix form the interior-node potentials $\mathbf{U}$ and temperatures $\mathbf{T}$ satisfy
\begin{equation}
  \mathbf{G}\,\mathbf{U} \;=\; \mathbf{S}\,\mathbf{T},
  \label{eq:GU=ST}
\end{equation}
with $\mathbf{G}$ the conductance matrix and $\mathbf{S}$ the Seebeck-coefficient matrix. The boundary voltages are obtained from the solved node potentials and Seebeck sources via selection matrices $\mathbf{I}_U$, $\mathbf{I}_S$, and $\mathbf{I}_V$, \begin{equation} \mathbf{V} \;=\; \big(\mathbf{I}_U\,\mathbf{G}^{-1}\mathbf{S} +\mathbf{I}_S +\mathbf{I}_V\big)\,\mathbf{T} \;\triangleq\; \mathbf{A}\,\mathbf{T}, \label{eq:A-from-GS} \end{equation} where $\mathbf{I}_U$ selects the contributions of the interior-node potentials at the measured boundary channels, $\mathbf{I}_S$ selects the Seebeck contributions applied directly at the boundary channels, and $\mathbf{I}_V$ enforces the measurement reference used by the readout electronics (i.e., voltages are defined relative to a chosen datum). This reference is required because the network can be electrically floating, in which case the absolute potentials are only defined up to an additive constant; $\mathbf{I}_V$ fixes that gauge without changing physically meaningful voltage differences. The resulting boundary voltages were also verified by circuit-level simulation of the ThermoMesh network in LTspice. Together, Eqs.~\eqref{eq:boundary_map}–\eqref{eq:A-from-GS} define the ThermoMesh measurement model.

For each frame $n$, the \emph{spatial sparsity} of the temperature signal is the number of nonzero entries in $\mathbf{T}[n]$,
\begin{equation}
  \|\mathbf{T}[n]\|_0
  \;\triangleq\;
  \#\big\{j : T_j[n]\neq 0\big\},
\end{equation}
where $j$ indexes the $MN$ junctions. Over an observation sequence, the \emph{temporal sparsity} is defined as
\begin{equation}
  q_t \;\triangleq\; \max_n \|\mathbf{T}[n]\|_0,
\end{equation}
i.e., the largest number of simultaneously active junctions in any single frame of that sequence; this is equivalently the largest number of simultaneous events in any frame of the sequence.

We define the \emph{event duration} $\tau_e$ as the time interval during which a heat source remains detectable somewhere within the sensing area, including motion, ending when it exits the measurement region or cools below detectability.

The sparse-event regime considered in this work assumes scale separation in both time and space. Let $\tau_m$ denote the measurement-window duration, $\tau_e$ the event duration defined above, and $\tau_s$ the pixel characteristic heat-transfer (thermal equilibration) time constant. Thin-film non-contact operation can achieve $\tau_s$ down to about $100~\mu\mathrm{s}$ in thermal-mass-limited designs~\cite{chen2021ultrafast}, whereas contact operation can achieve $\tau_s$ down to about $50~\mathrm{ns}$ in thermal-diffusion-limited designs~\cite{choi2007fabrication}. The temporal sparse-event regime assumed here therefore requires $\tau_m \gg \tau_e \gg \tau_s$. Spatially, we distinguish the total sensor area $A_s$, the pixel area $A_p$, and the event footprint area $A_e$. We assume $A_s \gg A_p$ and that a single event excites at most one pixel within a given frame. Equivalently, the event must not raise neighboring pixels above detectability. Under this assumption, multi-pixel excitation by a single localized event is not considered in the present work.

We define four performance metrics: \emph{range}, \emph{efficiency}, \emph{sensitivity}, and \emph{accuracy}.

\subsubsection{Range} \label{sec:rangePer}
In this work, \emph{range} refers to the operating regimes over which the sensor and the associated reconstruction are specified and characterized. In addition to the conventional temperature operating range, we also characterize a \emph{sparsity range}, which is specific to the proposed in-sensor compression framework.

The \emph{temperature range} is the interval $[T_{\min},T_{\max}]$ of junction temperatures over which the sensor and reconstruction are specified and characterized.

For a given design and reconstruction method, the sensor is \emph{characterized up to spatial sparsity} $q_s^{\max}$ if its performance has been characterized for all frames satisfying $\|\mathbf{T}[n]\|_0\le q_s^{\max}$. The set
\begin{equation}
  \{0,1,\ldots,q_s^{\max}\}
\end{equation}
is then the \emph{spatial sparsity range} of the sensor.

Likewise, the sensor is \emph{characterized up to temporal sparsity} $q_t^{\max}$ if its performance has been characterized for all observation sequences satisfying $q_t \le q_t^{\max}$. The set
\begin{equation}
  \{0,1,\ldots,q_t^{\max}\}
\end{equation}
is then the \emph{temporal sparsity range} of the sensor.

In this work we focus on $q_s^{\max}=1$ and $q_t^{\max}=1$.

\subsubsection{Efficiency}

\emph{Efficiency} measures the dynamic readout energy per frame. Let $E_s$ denote the energy required by the readout electronics (e.g., analog front-end and ADC/DAQ) to acquire one boundary-channel voltage sample. Then the ADC energy per frame scales with the number of readout channels:
\begin{equation}
E_{\text{frame}} \;\propto\; N_{\text{read}}\,E_s .
\end{equation}

Let $N_{\text{pix}}$ denote the total number of pixels and $N_{\text{read}}$ denote the number of readout channels. A conventional imager reads all pixels and therefore uses $N_{\text{pix}}$ readout channels, whereas ThermoMesh uses only $N_{\text{read}}$ boundary channels. The resulting channel-count reduction factor is
\begin{equation}
\eta_{\text{ch}} \;=\; \frac{N_{\text{pix}}}{N_{\text{read}}}.
\end{equation}

Because ThermoMesh uses perimeter-only readout, $N_{\text{read}}$ scales with the linear dimension of the array. Equivalently, Eq.~(14) scales linearly with $\sqrt{N_{\text{pix}}}$, i.e., with the square root of the number of pixels assuming a square sensing area.

\subsubsection{Sensitivity}
\emph{Sensitivity} is characterized by the boundary–voltage swing generated by a unit temperature increase at interior junction $j$,
\begin{equation}
  \sigma_j \;=\; \max_i A_{ij} \;-\; \min_i A_{ij} \quad [\mathrm{V/K}],
  \label{eq:sensitivity-node}
\end{equation}
and by the minimum sensitivity across the array,
\begin{equation}
  \sigma_{\min} \;=\; \min_j \sigma_j.
  \label{eq:sensitivity-min}
\end{equation}

To quantify nonlinearity we follow the log--log slope metric used for super-linear sensor response~\cite{wan2023sensor}, where super-linearity is defined with respect to output magnitude and stimulus intensity. In the nonlinear-interlayer case, the sensing matrix depends on temperature, so the boundary response depends on the stimulus level. We therefore fix the ambient temperature at $T_{\mathrm{amb}}$ and heat a specific junction (the center) by an amount $\Delta T$ above the ambient temperature. Let $\left.\Delta V_{\mathrm{center}}(\Delta T)\right|_{T=T_{\mathrm{amb}}}$ denote the resulting boundary-voltage swing. Following the log--log slope metric used for super-linear sensor response~\cite{wan2023sensor}, we define the super-linearity exponent
\begin{equation}
  \left.\kappa(\Delta T)\right|_{T=T_{\mathrm{amb}}}
  \;=\;
  \frac{d\ln \left.\Delta V_{\mathrm{center}}(\Delta T)\right|_{T=T_{\mathrm{amb}}}}{d\ln \Delta T}
  \label{eq:kappa_fromR}
\end{equation}
so that $\kappa>1$ indicates a super-linear increase of boundary response with stimulus amplitude.

\subsubsection{Accuracy}\label{subsubsec:accuracy-metrics}
Accuracy is evaluated at two levels. At the task level, heat-source \emph{localization} is treated as a \emph{classification} problem (one class per pixel), and temperature \emph{estimation} is treated as a \emph{regression} problem.

For localization, each pixel is assigned a unique class index $c\in\{1,\ldots,MN\}$ using the same indexing as in the dataset. Let $N_{\mathrm{eval}}$ denote the number of evaluation samples, and let $c_n$ and $\hat{c}_n$ denote the ground-truth and predicted class indices for sample $n\in\{1,\ldots,N_{\mathrm{eval}}\}$. Localization accuracy is defined as
\begin{equation}
\mathrm{Acc}
\,=\,
\frac{1}{N_{\mathrm{eval}}}
\sum_{n=1}^{N_{\mathrm{eval}}}
\mathbf{1}\!\left\{\hat{c}_n=c_n\right\},
\end{equation}
where $\mathbf{1}\{\cdot\}$ is the indicator function.

To quantify localization errors when $\hat{c}_n\neq c_n$, we map each class index $c$ to its 2D pixel location $\boldsymbol{\phi}(c)=(x_c,y_c)$ on the $M\times N$ grid using the same class-to-location mapping as in the dataset. The (per-sample) normalized spatial error is
\begin{equation}
d_{\rm norm}(n)=
\frac{\left\|\,\boldsymbol{\phi}\!\left(\hat{c}_n\right)
-\boldsymbol{\phi}\!\left(c_n\right)\,\right\|_2}
{\sqrt{(M-1)^2+(N-1)^2}},
\label{eq:dnorm}
\end{equation}
i.e., Euclidean pixel distance normalized by the mesh diagonal.

For temperature estimation, let $T_n$ and $\hat{T}_n$ denote the ground-truth and predicted temperatures for sample $n$. The mean absolute error (MAE) is
\begin{equation}
  \mathrm{MAE}
  \;=\;
  \frac{1}{N_{\mathrm{eval}}}
  \sum_{n=1}^{N_{\mathrm{eval}}}
  \left|\hat{T}_n - T_n\right|.
\end{equation}

For the device-level evaluation, we report the noise-equivalent temperature (NET),
\begin{equation}
  \mathrm{NET}
  \;\approx\;
  \frac{\mathrm{std}\!\big(\boldsymbol{\nu}\big)}{\big\|\,\Delta\mathbf{V}/\Delta T\,\big\|},
  \label{eq:net-def}
\end{equation}
where $\boldsymbol{\nu}$ is the boundary-voltage noise vector, $\mathrm{std}(\boldsymbol{\nu})$ denotes the standard deviation computed across the boundary channels, and $\|\,\Delta\mathbf{V}/\Delta T\,\|$ is an aggregate sensitivity. NET summarizes the temperature change that produces a boundary-voltage change comparable to the boundary-noise level.

\subsection{An Analog--Digital Thin-Film Thermal Sensor for 2D Heat-Source Localization and Strength Estimation}

We apply the ThermoMesh measurement model to the \emph{single-pixel, single-event} operating regime ($q_s^{\max}=1$, $q_t^{\max}=1$) used throughout this paper. Consider a single discrete frame $n$ and the corresponding junction-temperature vector $\mathbf{T}[n]\in\mathbb{R}^{MN}$ defined on the $M\times N$ grid. A \emph{single localized heating event} in this framework means that, at that frame, at most one sensing junction (equivalently, one pixel) exhibits a nonzero temperature rise relative to the reference, i.e.,
\begin{equation}
\mathbf{T}[n]\in\mathbb{R}^{MN},\qquad
T_{r^{\ast},c^{\ast}}[n]=\Delta T[n],\qquad
T_{i,j}[n]=0 \;\text{ for } (i,j)\neq(r^{\ast},c^{\ast}),
\label{eq:single_pixel_T}
\end{equation}
where $(r^{\ast},c^{\ast})$ denotes the row and column indices of the active (heated) junction and $\Delta T[n]$ is its amplitude on the junction grid. Physically, under the sparse-event scale separations introduced in Section~\ref{sec:design-metrics}, this regime corresponds to sufficiently localized heating such that neighboring junctions do not show a noticeable temperature rise within the same frame.

\begin{figure}[!t]
  \centering
  \includegraphics[width=\linewidth,height=\textheight,keepaspectratio]{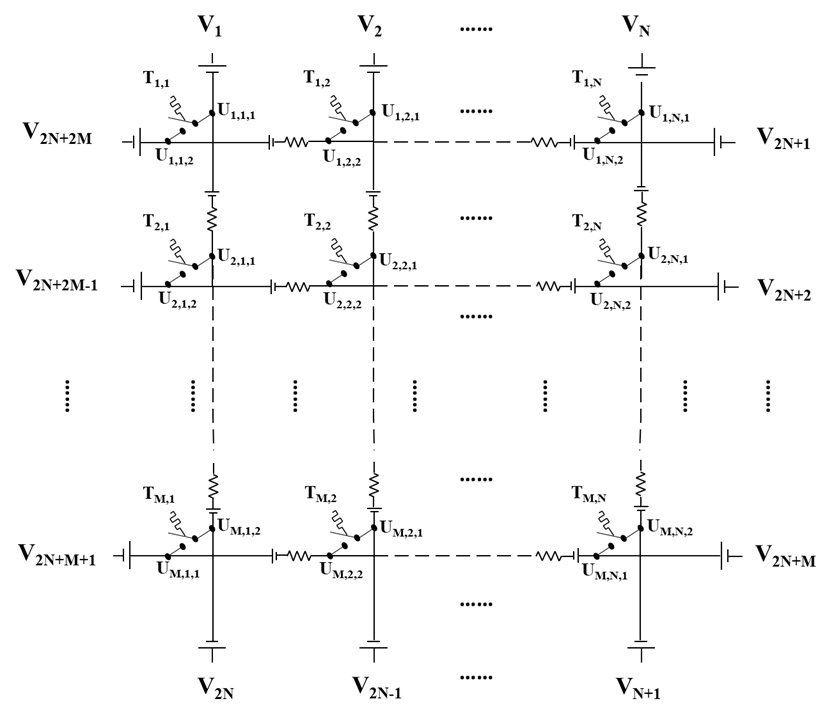}
  \caption{Ideal-switch equivalent circuit.}
  \label{fig:ideal-switch-circuit}
\end{figure}

In the idealized temperature-dependent interlayer limit, each crossing behaves as a \emph{thermal switch} with resistance
\begin{equation}
  R(T) \;=\;
  \begin{cases}
    0,      & T \ge T_{\mathrm{th}},\\[2pt]
    \infty, & T <  T_{\mathrm{th}},
  \end{cases}
  \label{eq:thermal-switch}
\end{equation}
so only the heated junction $(r^{\ast},c^{\ast})$ closes while all other crossings remain open. Here $T_{\mathrm{th}}$ is the switching threshold at which the interlayer transitions from the high-resistance (open) state to the low-resistance (closed) state. Under this ideal-switch behavior, the boundary measurement vector $\mathbf{V}[n]$ contains a \emph{digital} location signature and an \emph{analog} magnitude: the boundary channels that are electrically connected to the closed row and column exhibit defined voltages relative to the measurement reference, and their indices identify the active row and column (and thus the 2D location $(r^{\ast},c^{\ast})$), while their magnitudes depend on $\Delta T[n]$ through the effective Seebeck coefficients and the network conductances. Boundary channels with no conductive path to the closed junction are electrically floating (open-circuit) under this ideal model and do not carry location information. An equivalent-circuit schematic---showing the boundary channels, internal node potentials, and junction temperatures---is provided in Fig.~\ref{fig:ideal-switch-circuit}.

A perfect thermal switch that satisfies all design constraints is not always available in practice. Accordingly, the remainder of the paper analyzes three interlayer realizations introduced in Fig.~\ref{fig:tm-variants}: (i) \emph{baseline (no interlayer)}~\cite{zhao2019thermomesh}, in which only the native thermocouple junction is present; (ii) a \emph{constant-resistance interlayer (linear)}, where the interlayer behaves as a temperature-independent resistor; and (iii) a \emph{temperature-dependent interlayer (nonlinear)}, where the interlayer resistance depends on temperature.

\subsection{ThermoMesh with Linear Resistance Interlayer}

\begin{figure}[!t]
  \centering
  \includegraphics[width=\linewidth,height=\textheight,keepaspectratio]{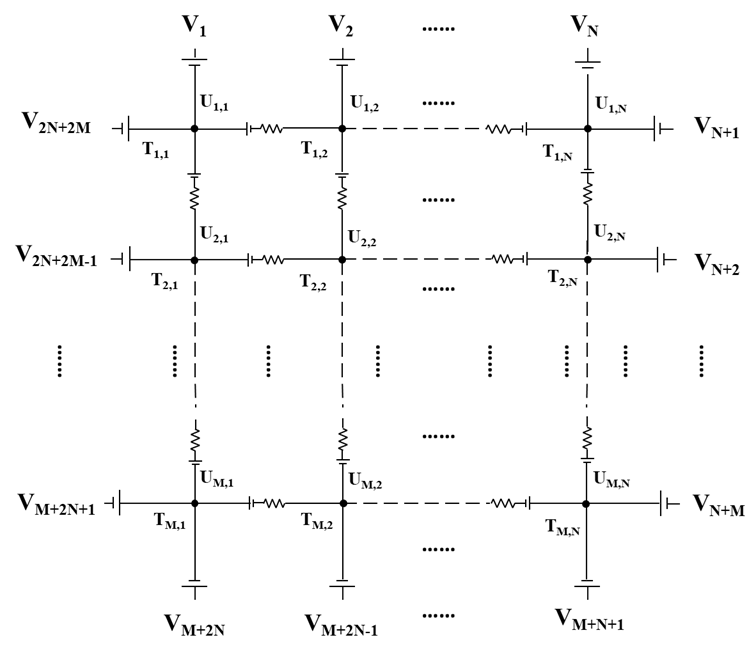}
  \caption{Equivalent circuit used to represent the baseline ThermoMesh (no interlayer).}
  \label{fig:circ-baseline}
\end{figure}

We begin from the baseline ThermoMesh containing only crossed thermocouple wires and no resistive interlayer. Its 2D equivalent circuit, used to assemble the linear measurement model in Eq.~\eqref{eq:boundary_map}, is shown in Fig.~\ref{fig:circ-baseline}. In this configuration the boundary channels are measured to form the voltage vector $\mathbf{V}$; interior temperatures are reconstructed from the boundary measurements.

\begin{figure}[!t]
  \centering
  \includegraphics[width=\linewidth,height=\textheight,keepaspectratio]{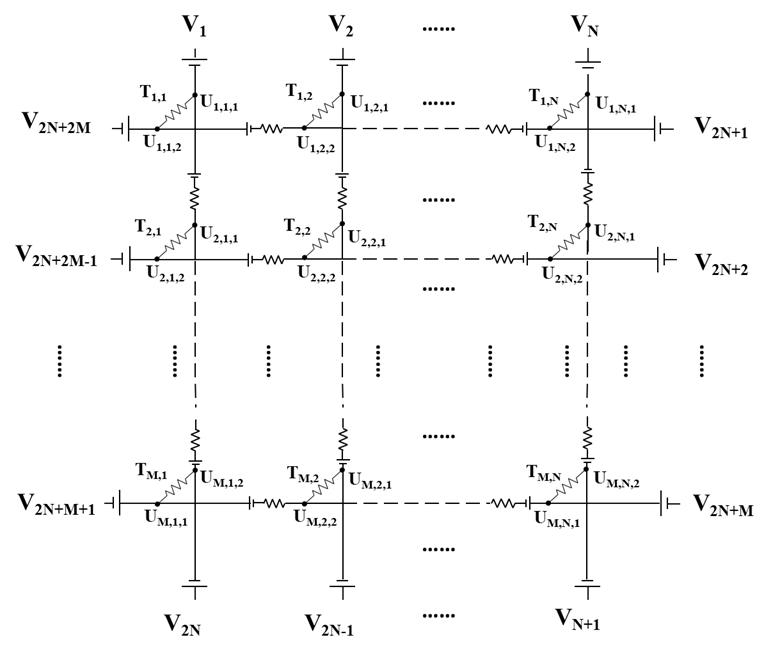}
  \caption{Equivalent circuit of ThermoMesh with a linear resistive interlayer.}
  \label{fig:circ-linear}
\end{figure}

The baseline sensitivity is strongly edge-biased. The weakest region (near the center) has sensitivity well below the K-type thermocouple typical value ($\sim\!40~\mu$V/K), which limits detection near the middle of the mesh. To mitigate this while preserving the linear model in Eq.~\eqref{eq:boundary_map}, we introduce a thin \emph{linear resistive interlayer} between the stacked films. Electrically, this adds temperature-independent shunt paths at each crossing and reshapes the sensitivity matrix while keeping the model linear. The modified circuit used in subsequent calculations is shown in Fig.~\ref{fig:circ-linear}. In the linear case, the temperature range is set primarily by the thermocouple material rather than the interlayer; for K-type thermocouples, it is on the order of $1600$~K.

Uniqueness of sparse recovery in the linear-interlayer case is assessed via the Null Space Property (NSP): if the sensitivity matrix $\mathbf{A}$ satisfies the NSP of order $q_s^{\max}$, then any temperature field $\mathbf{T}$ with $\|\mathbf{T}\|_0 \le q_s^{\max}$ is uniquely determined by the measured boundary voltages $\mathbf{V}$. We verified that the sensitivity matrices $\mathbf{A}$ assembled from the network satisfy the NSP for $q_s^{\max}=1$ across multiple array sizes (up to $200\times200$). Therefore, in the linear case, a $1$-sparse temperature field is uniquely determined by the boundary measurements, and this is consistent with the Orthogonal Matching Pursuit (OMP) recovery results reported later in this subsection.

Per-junction sensitivity $\sigma_j$ and minimum sensitivity $\sigma_{\min}$ are computed using the definitions given once in Section~\ref{sec:design-metrics} (Eqs.~\eqref{eq:sensitivity-node}–\eqref{eq:sensitivity-min}); because the interlayer is temperature independent, \(\sigma_j\) and \(\sigma_{\min}\) are temperature independent for a fixed geometry and materials set.

\begin{figure}[!t]
    \centering

    \includegraphics[width=\linewidth,height=0.36\textheight,keepaspectratio]{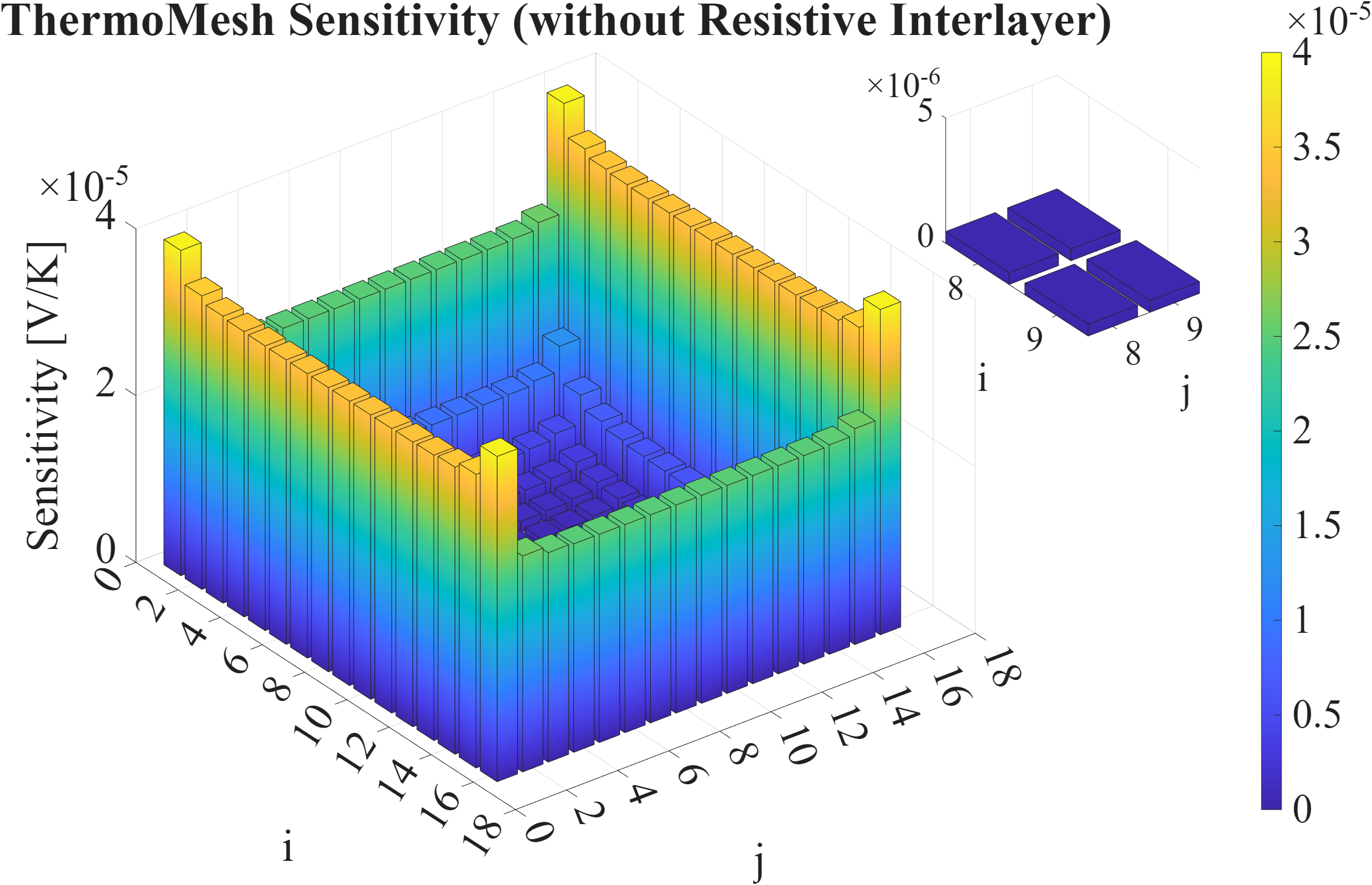}
    \par\smallskip\textbf{(a)}

    \vspace{5mm}

    \includegraphics[width=\linewidth,height=0.36\textheight,keepaspectratio]{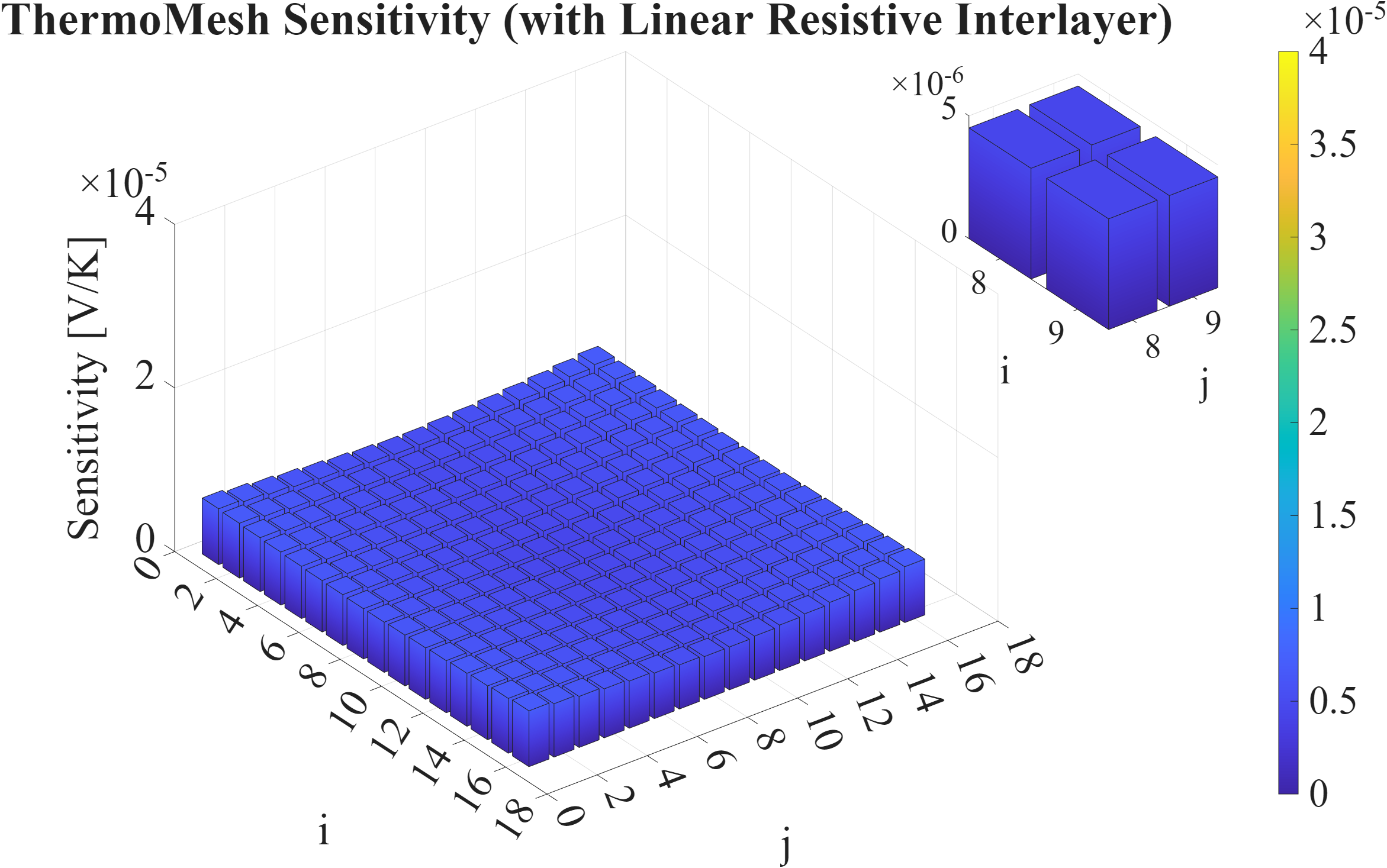}
    \par\smallskip\textbf{(b)}

    \caption{\textbf{Sensitivity maps for a 16$\times$16 ThermoMesh.}
    (a) Baseline configuration without a resistive interlayer, showing strong edge-dominated sensitivity and reduced response near the center.
    (b) Corresponding map with a linear resistive interlayer, illustrating redistribution and flattening of sensitivity across the mesh.}
    \label{fig:sens-compare}
\end{figure}

The effect of the linear interlayer on sensitivity is shown in Fig.~\ref{fig:sens-compare} by comparing the baseline map (no interlayer) with the corresponding map obtained with the linear resistive interlayer. Panel~(a) shows the baseline distribution; panel~(b) shows the distribution with the linear interlayer. In both panels, the inset is a zoom of the central junctions using the \emph{same scale and colorbar} to enable a direct comparison. With the interlayer, the minimum (central-junction) sensitivity rises from $4.67\times10^{-7}$~V/K to $4.51\times10^{-6}$~V/K (a $\approx 10\times$ increase), and the profile flattens markedly. Although junctions near the boundary become less sensitive than in the baseline, the minimum sensitivity increases and the spatial distribution of sensitivity becomes substantially more uniform overall.

\begin{figure}[!t]
  \centering
  \includegraphics[width=\linewidth,height=\textheight,keepaspectratio]{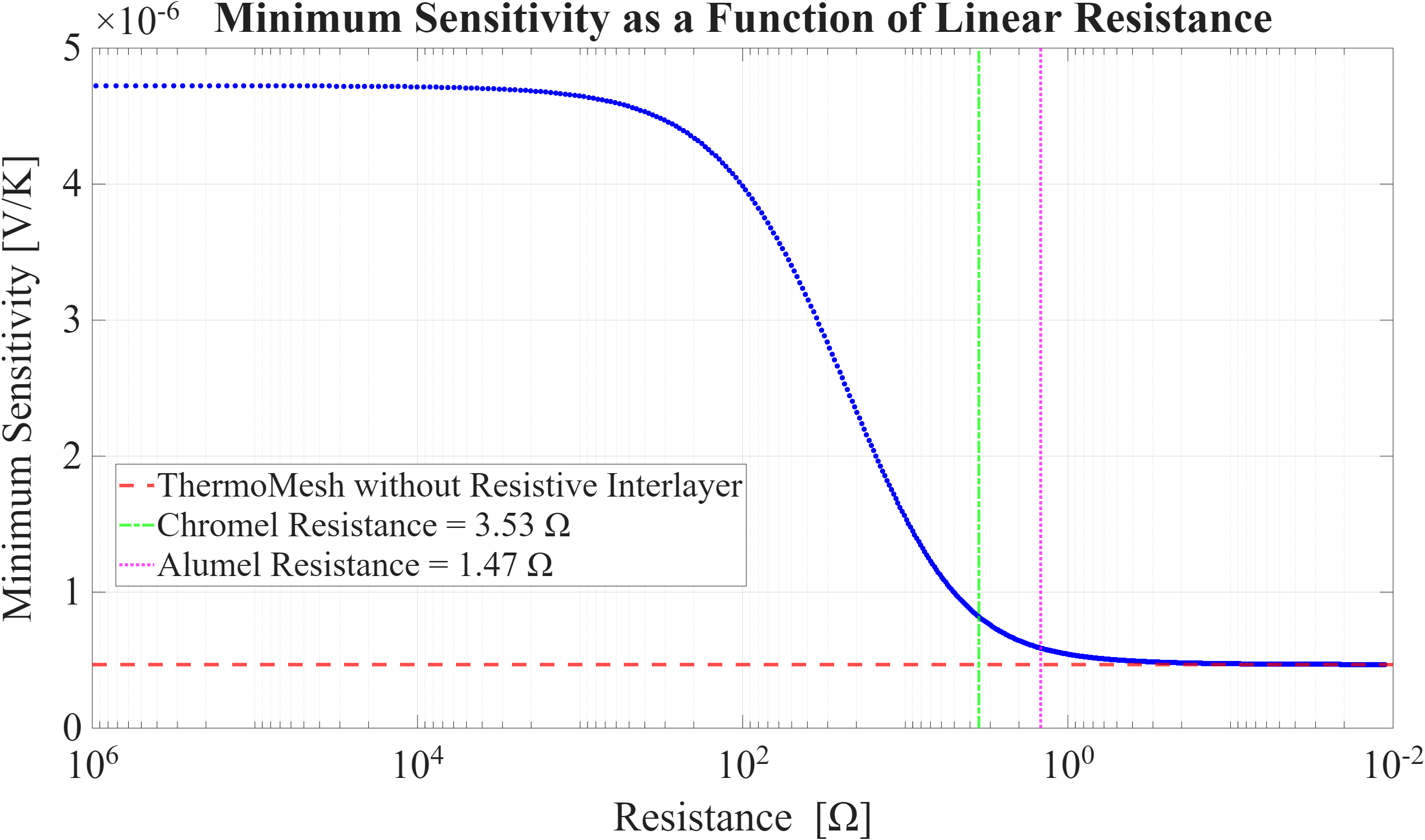}
  \caption{Minimum sensitivity versus linear interlayer resistance for a $16\times16$ mesh. The dashed red line denotes the baseline (no interlayer). As $R$ increases, $\sigma_{\min}$ increases and approaches a high-$R$ plateau; as $R\!\to\!0$, the curve returns to the baseline. Vertical dashed lines mark representative Chromel and Alumel resistances for reference.}
  \label{fig:min-sens-R-16}
\end{figure}

Treating the interlayer resistance $R$ as a design parameter shows that the minimum sensitivity depends strongly on $R$. Fig.~\ref{fig:min-sens-R-16} plots $\sigma_{\min}$ for a $16\times16$ mesh: as $R$ increases, $\sigma_{\min}$ increases relative to the no-interlayer baseline and then changes more slowly at larger $R$. As $R\!\to\!0$, the curve approaches the baseline (dashed red), which matches the no-interlayer case. The vertical markers indicate representative resistances associated with the thermocouple legs (Chromel and Alumel), showing where practical values fall relative to the transition region.

\begin{figure}[!t]
  \centering
  \includegraphics[width=\linewidth,height=\textheight,keepaspectratio]{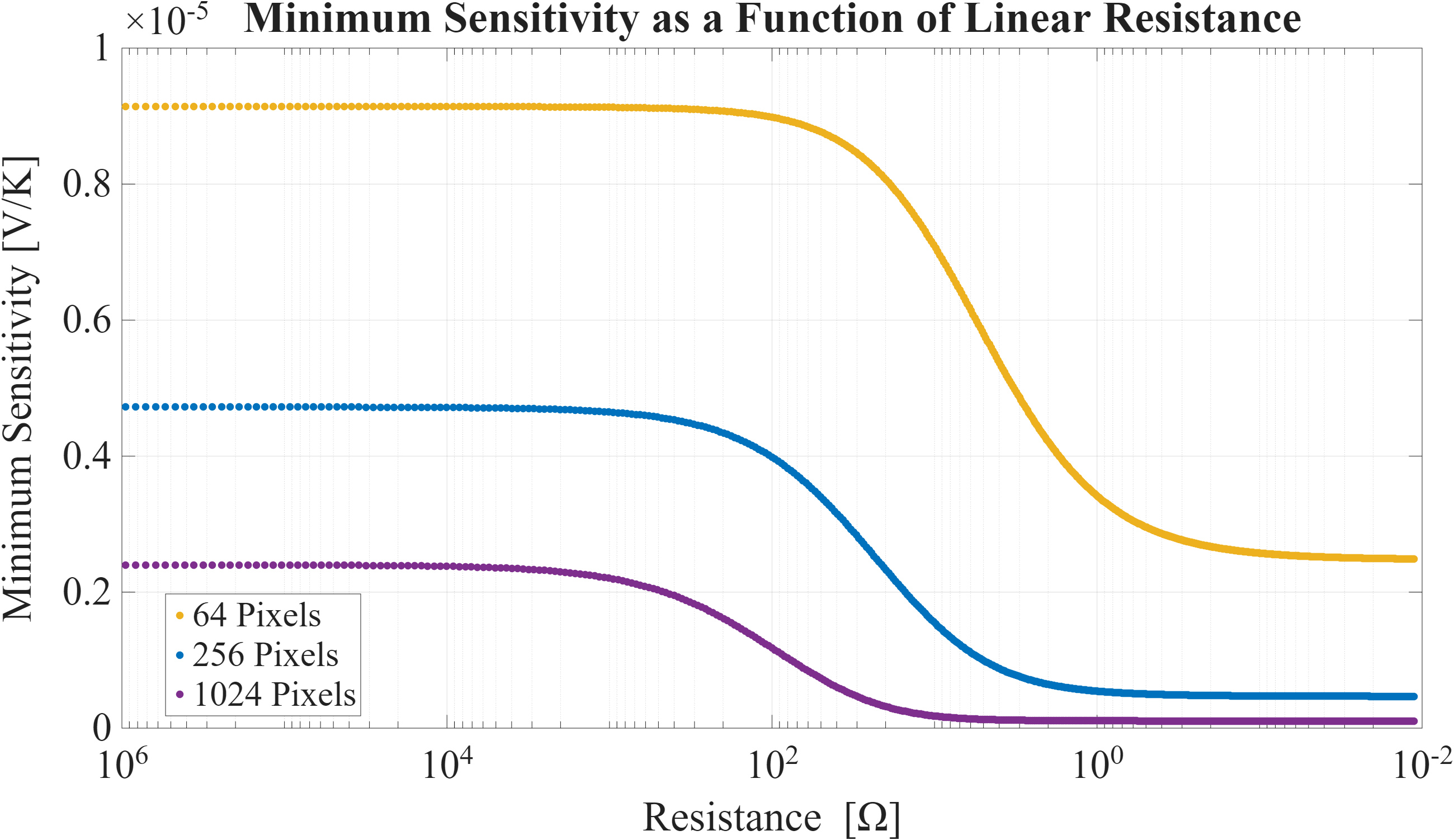}
  \caption{Minimum sensitivity versus linear interlayer resistance across mesh sizes. For each size, $\sigma_{\min}$ increases with $R$ and saturates at a high-$R$ plateau, while the absolute plateau decreases with increasing resolution.}
  \label{fig:min-sens-R-sizes}
\end{figure}

Scaling the mesh shows two consistent trends (Fig.~\ref{fig:min-sens-R-sizes}). First, for each size the minimum sensitivity increases with $R$ and then changes more slowly at higher $R$, with the transition occurring over an intermediate-$R$ range. Second, the high-$R$ level decreases as resolution increases, reflecting the sensitivity--resolution trade-off under boundary-only readout. Relative to the baseline, the improvement provided by the resistive interlayer increases with array size; the corresponding improvement factors are summarized in Table~\ref{tab:improve-factors}.

\begin{table}[t]
  \centering
  \caption{Minimum sensitivity improvement (relative to baseline) for the linear-interlayer design.}
  \label{tab:improve-factors}
  \begin{tabular}{cc}
    \toprule
    \textbf{Mesh size ($M\times N$ pixels)} & \textbf{Minimum Sensitivity Improvement} \\
    \midrule
    64   & $4\times$ \\
    256  & $10\times$ \\
    1024 & $23\times$ \\
    \bottomrule
  \end{tabular}
\end{table}

To assess robustness, we formed a noise-free boundary signal $\mathbf{V}_{\text{noise-free}}=\mathbf{A}\mathbf{T}$, computed its mean–square power $P_s$, chose a target signal-to-noise ratio ($\mathrm{SNR_{dB}}$), and set the noise power to $P_n=P_s/10^{\mathrm{SNR_{dB}}/10}$. We then drew a white Gaussian noise vector $\boldsymbol{\nu}\sim\mathcal{N}(\mathbf{0},P_n\mathbf{I})$ and formed the noisy measurements $\mathbf{V}=\mathbf{V}_{\text{noise-free}}+\boldsymbol{\nu}$. This models aggregate electronic/thermal disturbances as independent, zero-mean noise at the boundary channels with variance fixed by the desired SNR. Recovery was performed with OMP; success was recorded when the active junction and its amplitude were recovered. The quantitative results for 16$\times$16 meshgrid and $q_s^{\max}=1$ are listed in Table~\ref{tab:omp}. Here, our goal is to illustrate noise effects; in future work we will incorporate realistic sources such as Johnson–Nyquist noise and electromagnetic interference (EMI) into the sensor-level model.
\begin{table}[t]
  \centering
  \caption{Recovery success via OMP for the linear-interlayer design.}
  \label{tab:omp}
  \begin{tabular}{cc}
    \toprule
    \textbf{SNR (dB)} & \textbf{Recovery success rate (\%)} \\
    \midrule
    No Noise & 100 \\
    40 & 73 \\
    20 & 12 \\
    \bottomrule
  \end{tabular}
\end{table}

Taken together, Figs.~\ref{fig:circ-baseline}–\ref{fig:min-sens-R-sizes} and Tables~\ref{tab:improve-factors}–\ref{tab:omp} show that a constant-resistance interlayer improves the minimum sensitivity and uniformity for a ThermoMesh, yet the absolute minimum sensitivity still declines with resolution. This motivates the nonlinear (temperature-dependent) interlayer studied next.

\subsection{ThermoMesh with Nonlinear Resistance Interlayer}
\label{sec:nonlinear}

\begin{figure}[!t]
  \centering
  \includegraphics[width=\linewidth,height=\textheight,keepaspectratio]{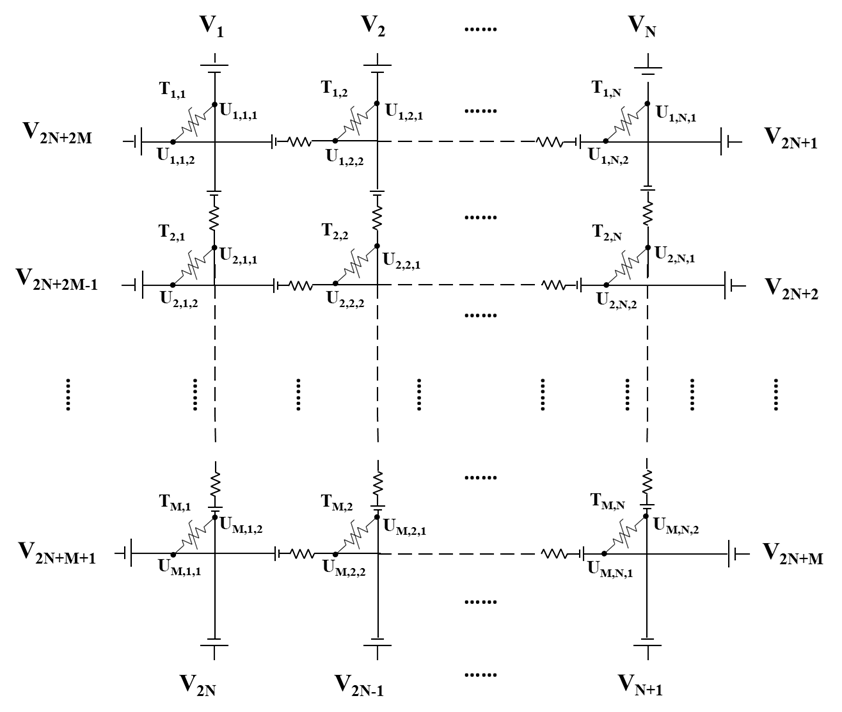}
  \caption{Equivalent circuit of ThermoMesh with a thermistor at each crossing (the interlayer).}
  \label{fig:nonlinear-overview}
\end{figure}

A practical realization of the ``thermal switch'' in Fig.~\ref{fig:ideal-switch-circuit} is a \emph{thermistor} placed between the orthogonal thermocouple films. Its resistance varies strongly with temperature but does not exhibit an ideal step change. As a result, the boundary-voltage vector is related to the junction-temperature vector through the nonlinear relation in Eq.~\eqref{eq:nonlinear-forward}, where the temperature-dependent matrix $\mathbf{A}(\mathbf{T})$ is obtained by enforcing Kirchhoff’s current law on the network. The corresponding equivalent circuit is shown in Fig.~\ref{fig:nonlinear-overview}.

For NTC thin films we parameterize the resistivity with the standard model,
\begin{equation}
  \rho(T)=\rho_0\,\exp\!\Big[\beta\Big(\tfrac{1}{T}-\tfrac{1}{T_0}\Big)\Big],
  \label{eq:beta}
\end{equation}
using different $(\rho_0,\beta,T_0)$ sets appropriate to each material system. Here, $\rho(T)$ is the interlayer \emph{resistivity}, $\rho_0$ is the resistivity at the reference temperature $T_0$, and $\beta$ is the thermistor coefficient. $T$ and $T_0$ are absolute temperatures in kelvin; for NTC materials, $\beta>0$, so $\rho(T)$ decreases as $T$ increases. For high-temperature operation we use perovskite thermistor ceramics~\cite{houivet2004high}. For lower-temperature operation (e.g., bolometers), we model VO$_2$ using a piecewise fit of Eq.~\eqref{eq:beta} across the metal–insulator transition (MIT) to capture the orders-of-magnitude resistivity change~\cite{ma2017influence}; recent material advances can further increase the transition sharpness and on/off ratio~\cite{cao2022enhancing}. In ThermoMesh the temperature range $[T_{\min},T_{\max}]$ is therefore set primarily by the interlayer material system rather than the mesh architecture; accordingly, we report results for a VO$_2$–NTC interlayer over $298$–$373 \mathrm{K}$ and for a ceramic-NTC interlayer over $973$–$1273 \mathrm{K}$.

We use the same sensitivity definitions as in the linear case (see Eqs.~\eqref{eq:sensitivity-node}–\eqref{eq:sensitivity-min}), with the important difference that both the per–junction sensitivity and the minimum sensitivity depend on the operating temperatures. For each nonlinear dataset, we therefore report two limits: the ambient minimum sensitivity \( \sigma_{\min}^{\mathrm{amb}} := \sigma_{\min}(\mathbf{T}_{\mathrm{amb}}) \) with all junctions at \(298\,\mathrm{K}\), and the event minimum sensitivity \( \sigma_{\min}^{\mathrm{hi}} := \sigma_{\min}(\mathbf{T}_{\mathrm{hi}}) \) with one hot junction at the application temperature while the others remain at \(298\,\mathrm{K}\). Also, to quantify super-linear behavior we use the super-linearity exponent defined in Eq.~\eqref{eq:kappa_fromR}, and values $\kappa>1$ indicate super-linear response.

\begin{figure}[!t]
  \centering
  \includegraphics[width=\linewidth,height=\textheight,keepaspectratio]{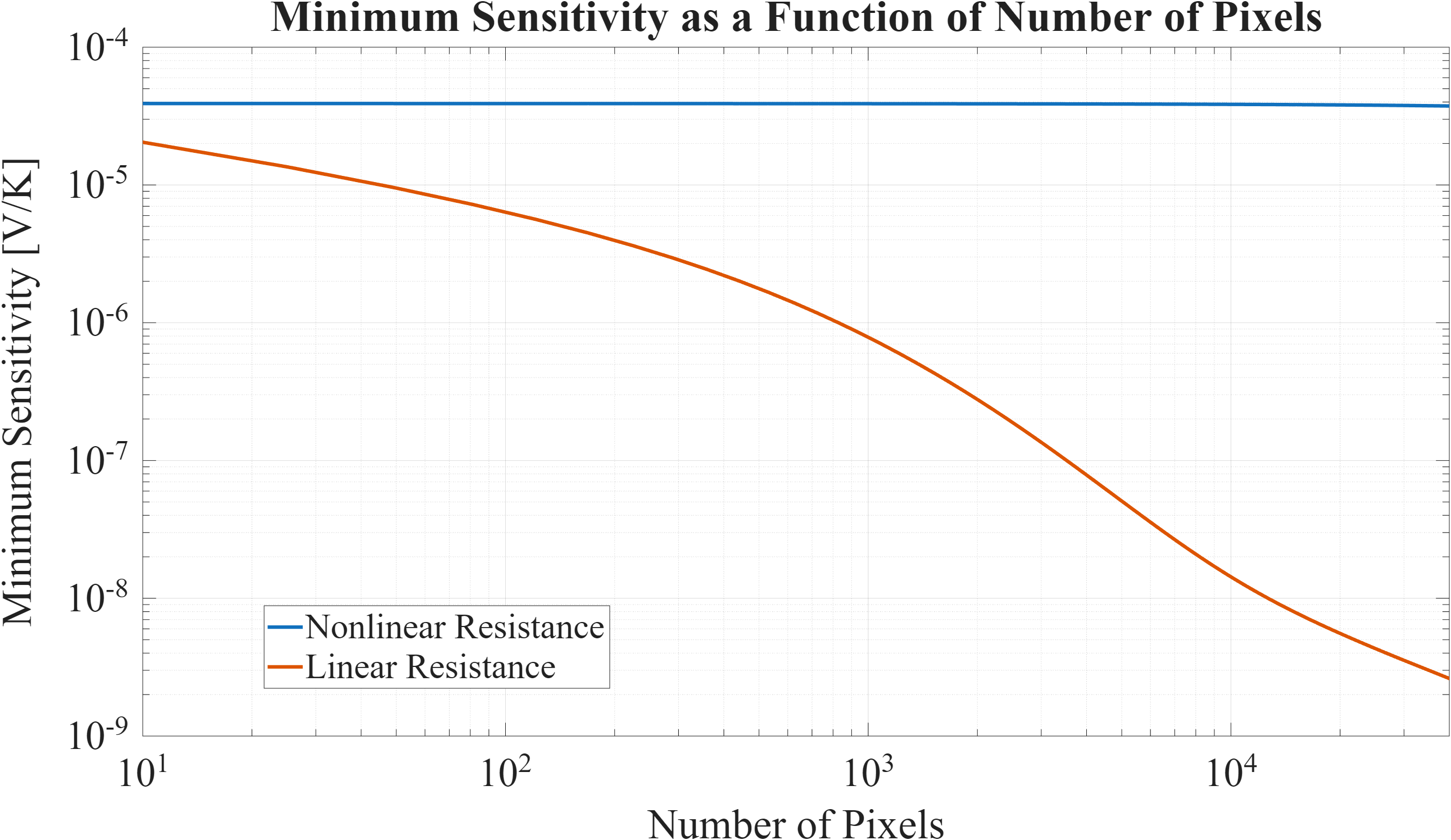}
  \caption{High-temperature case: minimum sensitivity vs.\ number of pixels using a ceramic-NTC interlayer compared with a linear interlayer.}
  \label{fig:minsens-ht}
\end{figure}

We first consider a perovskite-ceramic NTC layer for extreme events such as high-temperature molten droplets: the central junction is set to $1273\,\mathrm{K}$ while all other junctions remain at the ambient temperature. Fig.~\ref{fig:minsens-ht} plots the minimum sensitivity versus the number of pixels. The nonlinear interlayer maintains a higher minimum sensitivity as the array size increases, while the constant-$R$ case shows a much stronger decline. This is consistent with the thermal-switch interpretation: at the heated junction the interlayer resistance drops, which strengthens the coupling of the junction’s Seebeck response to the boundary readout. For the ceramic interlayer, the minimum sensitivity decreases from $3.90\times10^{-5}\,\mathrm{V/K}$ at $9$ pixels ($3\times3$) to $3.70\times10^{-5}\,\mathrm{V/K}$ at $40{,}000$ pixels ($200\times200$). For the linear case over the same scaling, $\sigma_{\min}$ drops from $2.14\times10^{-5}\,\mathrm{V/K}$ to $2.55\times10^{-9}\,\mathrm{V/K}$. Thus, for $200\times200$ meshgrid the nonlinear design achieves a $\sim 14{,}500\times$ higher minimum sensitivity.

\begin{figure}[!t]
  \centering
  \includegraphics[width=\linewidth,height=\textheight,keepaspectratio]{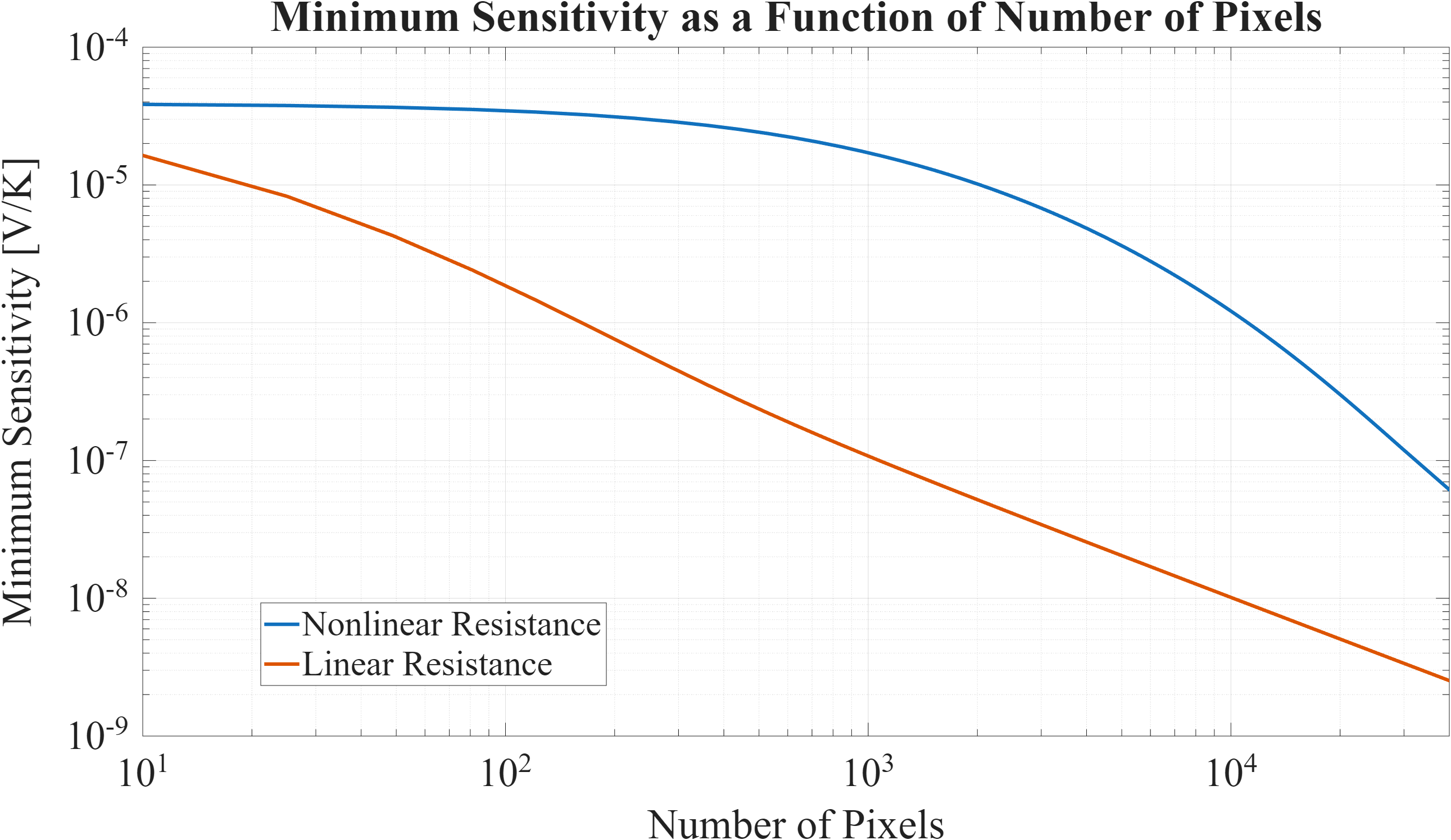}
  \caption{Low-temperature case: minimum sensitivity vs.\ number of pixels using a VO$_2$ interlayer compared with a linear interlayer.}
  \label{fig:minsens-lt}
\end{figure}

For applications with lower temperatures (e.g., microbolometers), we use a VO$_2$ interlayer and set the center to $373\,\mathrm{K}$ with all other junctions at the ambient temperature. Fig.~\ref{fig:minsens-lt} shows the minimum sensitivity versus pixel count for a VO$_2$ interlayer. For a VO$_2$ interlayer, the minimum sensitivity decreases from $3.85\times10^{-5}$ V/K at $9$ pixels to $6.00\times10^{-8}$ V/K at $40{,}000$ pixels, while the corresponding linear cases drop from $1.77\times10^{-5}$ V/K to $2.50\times10^{-9}$ V/K. For a $200\times200$ meshgrid, this design yields an $\sim24\times$ improvement in $\sigma_{\min}$ over the linear design.

To illustrate the nonlinear behavior explicitly, we show representative curves of minimum sensitivity and super-linearity for each nonlinear material. The VO$_2$ case exhibits a pronounced super-linear region across the MIT transition region reported in thin films~\cite{ma2017influence}, whereas the ceramic NTC case shows a milder but persistent super-linear response over the high-temperature range.
\begin{figure}[!t]
  \centering

  \begin{minipage}[t]{0.49\linewidth}
    \centering
    \includegraphics[width=\linewidth]{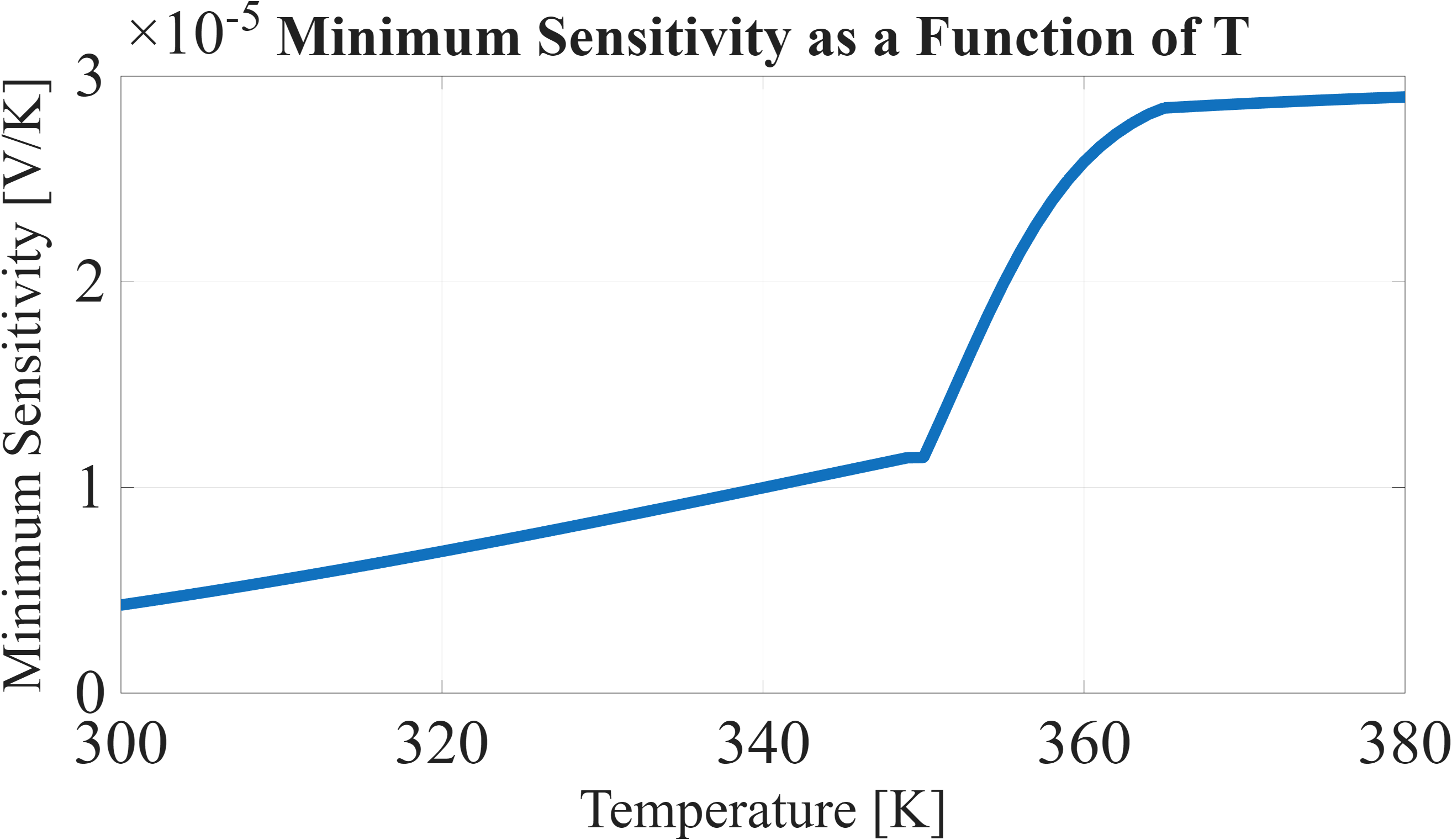}
    \par\smallskip\textbf{(a)}

    \vspace{4mm}

    \includegraphics[width=\linewidth]{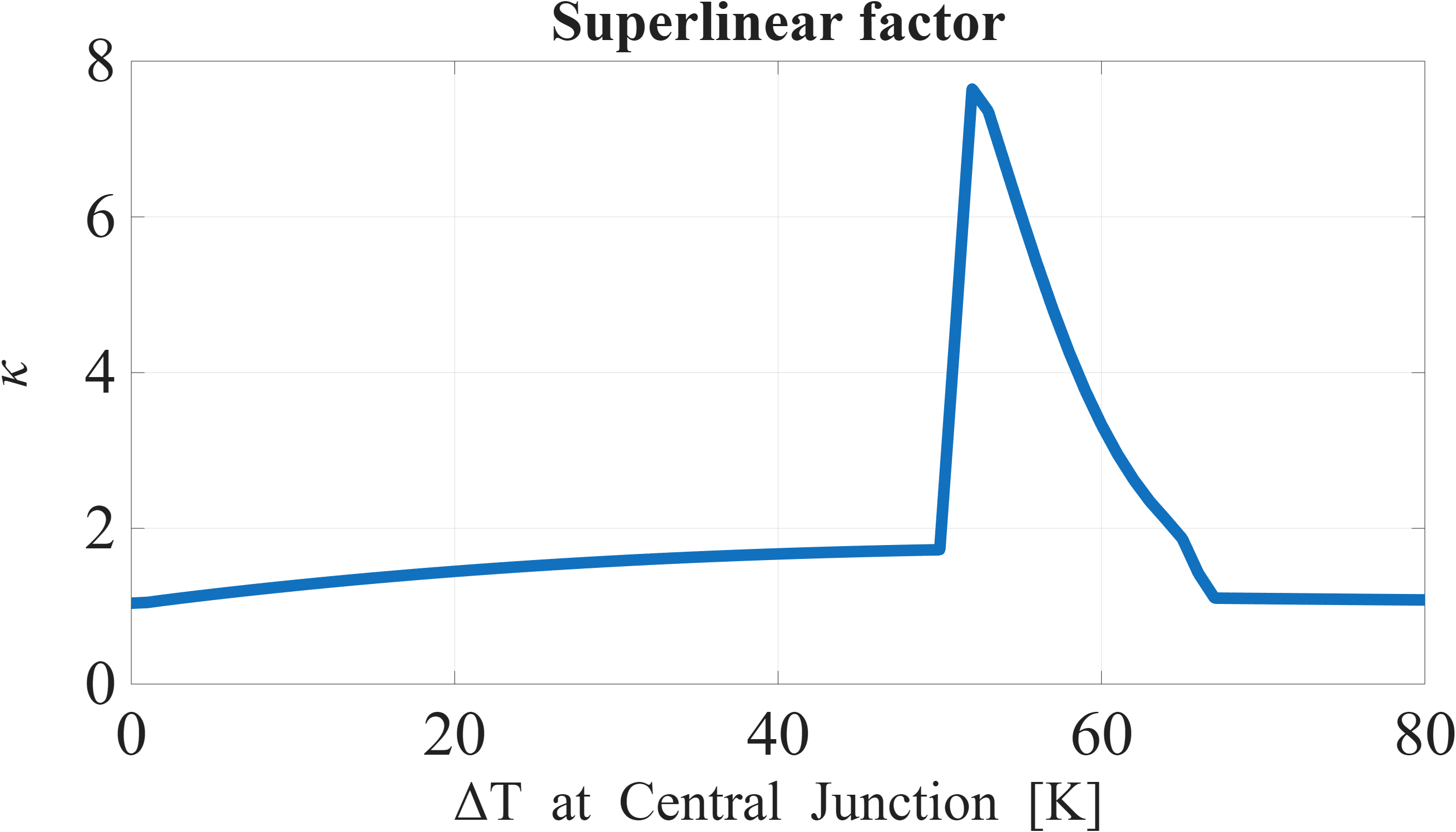}
    \par\smallskip\textbf{(c)}
  \end{minipage}\hfill
  \begin{minipage}[t]{0.49\linewidth}
    \centering
    \includegraphics[width=\linewidth]{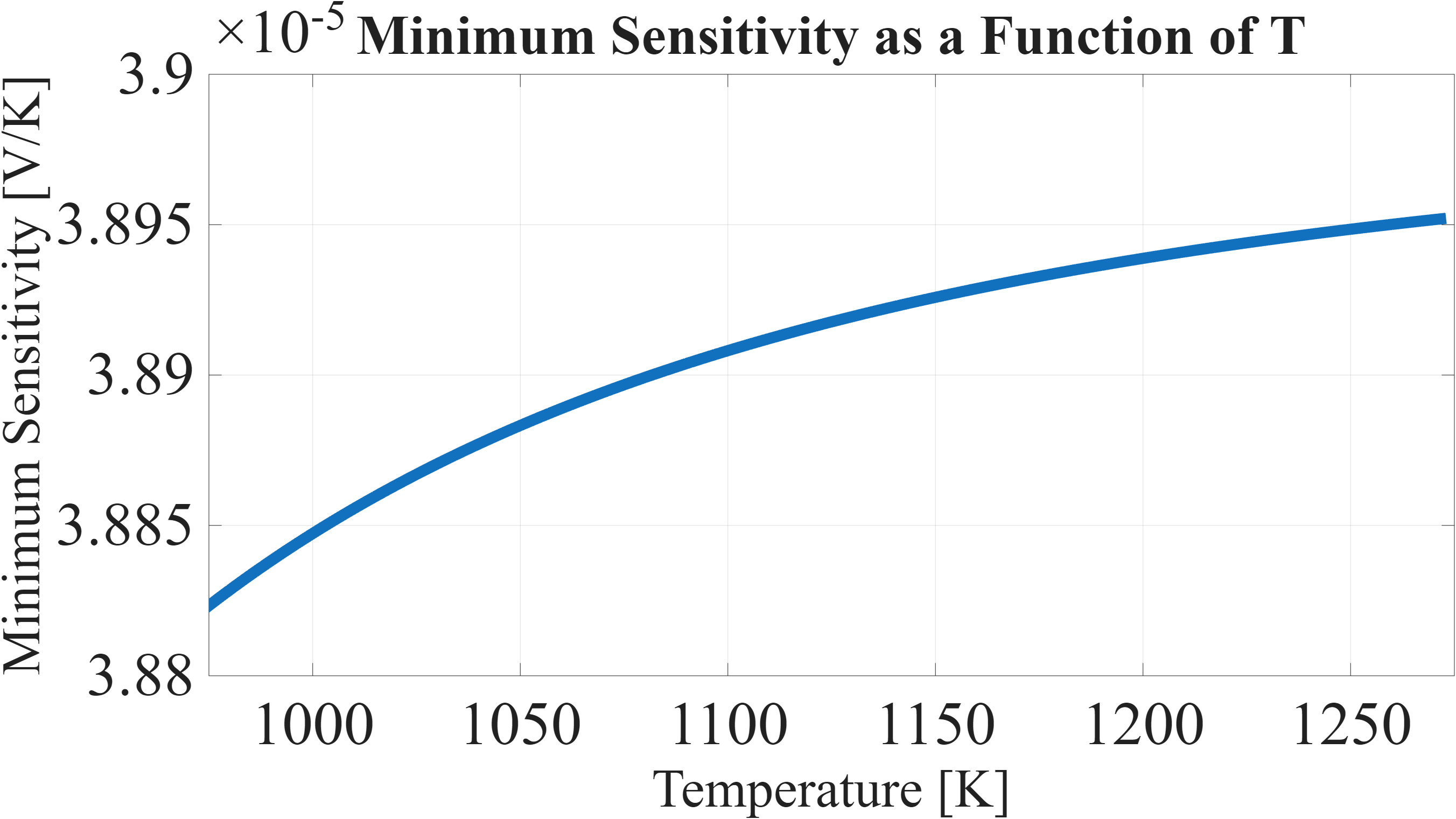}
    \par\smallskip\textbf{(b)}

    \vspace{4mm}

    \includegraphics[width=\linewidth]{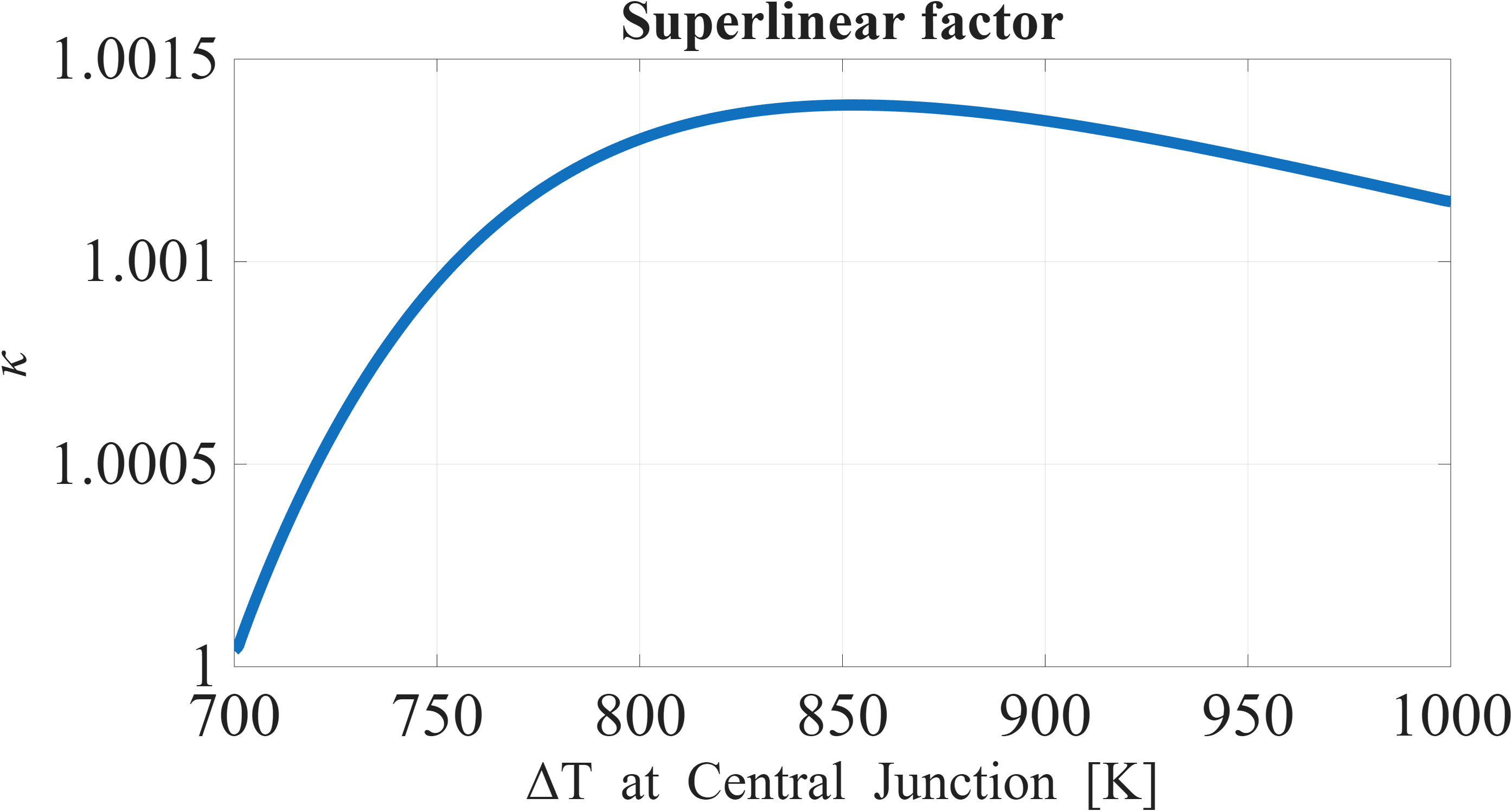}
    \par\smallskip\textbf{(d)}
  \end{minipage}

  \caption{\textbf{Temperature-dependent behavior of nonlinear interlayers.}
  (a--b) Minimum sensitivity vs.\ heated-junction temperature for vanadium-oxide (VO$_2$) and ceramic-NTC interlayers.
  (c--d) Super-linearity exponent $\left.\kappa(\Delta T)\right|_{T=T_{\mathrm{amb}}}$ vs.\ hot-spot temperature rise for the same interlayers.}
  \label{fig:minsens-vsT-combined}
\end{figure}

For the \(16\times16\) datasets underlying Fig.~\ref{fig:minsens-vsT-combined}, we report the two limits defined above: the ambient minimum sensitivity with all junctions at \(298\,\mathrm{K}\), and the event minimum sensitivity when only the central junction is heated to the application temperature. 
For VO\(_2\): \(\sigma_{\min}^{\mathrm{amb}}=4.06\times10^{-6}\,\mathrm{V/K}\) and \(\sigma_{\min}^{\mathrm{hi}}=2.91\times10^{-5}\,\mathrm{V/K}\) (center at \(383\,\mathrm{K}\)). 
For the ceramic NTC: \(\sigma_{\min}^{\mathrm{amb}}=4.45\times10^{-6}\,\mathrm{V/K}\) and \(\sigma_{\min}^{\mathrm{hi}}=3.895\times10^{-5}\,\mathrm{V/K}\) (center at \(1273\,\mathrm{K}\)). 
The corresponding super-linearity exponent $\left.\kappa(\Delta T)\right|_{T=T_{\mathrm{amb}}}$, extracted from the same sweeps with fixed ambient temperature, is shown in Fig.~\ref{fig:minsens-vsT-combined}(c--d).

We apply machine learning algorithms for heat-source localization and temperature estimation, following the ThermoNet approach of Zhao and Ye~\cite{zhao2019thermomesh}. Inputs are the boundary voltages $\mathbf V\in\mathbb{R}^{2M+2N}$; outputs are $MN$ classes (one per pixel). We use a modified long short-term memory (LSTM) network, we use a sequence input layer, a fully connected layer, a Softmax layer, and a classification layer (cross-entropy loss). The regressor maps to the temperature via a one-hidden-layer feedforward network (30 hidden units; default mean squared error (MSE) loss). We generate synthetic 1-sparse samples from the measurement model and use a fixed split consistent with the code: $60\%$ training, $40\%$ validation. For the classifier we use Adam with a mini-batch size of $256$ for $200$ epochs. 

We evaluate localization and temperature-estimation performance using the metrics defined in Section~\ref{subsubsec:accuracy-metrics}. Specifically, we report localization accuracy and temperature MAE across noise conditions, and we compute NET using~\eqref{eq:net-def}. Table~\ref{tab:ml-acc-net} reports localization accuracy and NET in the same format across interlayer options; Table~\ref{tab:ml-mae} reports temperature-regression MAE. At SNR $=20$~dB, the VO$_2$ case is slightly less accurate than the linear case despite its lower NET, which is attributed to the fact that the VO$_2$ sensitivity enhancement is concentrated near the MIT rather than uniformly across the full 298--373~K range.

To interpret localization errors, we additionally compute the normalized spatial error $d_{\rm norm}$ from Eq.~\eqref{eq:dnorm} for the misclassified test cases. Across our simulations, the distribution of $d_{\rm norm}$ for incorrect predictions concentrates below $d_{\rm norm}\le 0.2$, indicating that when the classifier is wrong, the predicted pixel is typically within $20\%$ of the mesh diagonal from the true heat-source location (i.e., errors are predominantly near-misses rather than large jumps).

\setlength{\tabcolsep}{3pt}


\begin{table}[t]
  \centering
  \caption{Localization accuracy and NET for interlayer options.}
  \label{tab:ml-acc-net}
  \small
  \setlength{\tabcolsep}{3pt}
  \renewcommand{\arraystretch}{1.15}
  \begin{tabularx}{\linewidth}{l c c c c X}
    \toprule
    \textbf{Resistance layer} &
    \textbf{Temp. range} &
    \textbf{No noise} &
    \textbf{40 dB} &
    \textbf{20 dB} &
    \textbf{NET range} \\
    \midrule
    Without resistance & 298--373 K  & 94\%  & 93\%  & 61\%  & 0.44--4.45 K \\
    Linear resistance  & 298--373 K  & 98\%  & 98\%  & 94\%  & 0.15--1.50 K\\
    NTC (VO$_2$)       & 298--373 K  & 98\%  & 98\%  & 91\%  & 0.07--0.72 K\\
    NTC (Ceramics)     & 973--1273 K & 100\% & 100\% & 100\% & 1.49--14.93 K \\
    \bottomrule
  \end{tabularx}
\end{table}



\begin{table}[t]
  \centering
  \caption{Temperature-estimation error measured by mean absolute error (MAE).}
  \label{tab:ml-mae}
  \small
  \setlength{\tabcolsep}{4pt}
  \renewcommand{\arraystretch}{1.15}
  \begin{tabular}{lcccc}
    \toprule
    \textbf{Resistance layer} &
    \textbf{Temp. range} &
    \textbf{No noise} &
    \textbf{40 dB} &
    \textbf{20 dB} \\
    \midrule
    Without resistance & 298--373 K  & 0.02 K & 0.48 K & 3.81 K \\
    Linear resistance  & 298--373 K  & 0.00 K & 0.16 K & 1.51 K \\
    NTC (VO$_2$)       & 298--373 K  & 0.01 K & 0.23 K & 1.60 K \\
    NTC (Ceramics)     & 973--1273 K & 0.01 K & 1.83 K & 16.72 K \\
    \bottomrule
  \end{tabular}
\end{table}

\section{Discussion}\label{sec:Discussion}

We situate ThermoMesh within the broader idea of analog computation as a physical mapping between stimuli and measurements. 
This work combines the analog computing method proposed by Kendall et al.~\cite{kendall2022scalable} with transduction mechanisms relying on phonon and electron transport. Related thermal analog-computing approaches based on heat transport have also been demonstrated using inverse-designed thermal metastructures that implement linear mappings via phonon conduction~\cite{silva2026thermal}. In the linear and nonlinear measurement models developed in Section~\ref{sec:Design and Performance}, ThermoMesh physically implements a map from a spatio–temporally sparse temperature field \( \mathbf{T} \) to boundary voltages \( \mathbf{V} \) via Seebeck transduction and an interlayer (constant \(R\) or NTC). In this sense, the mesh performs in-sensor analog computation. Interpreting the mesh as an analog computing substrate clarifies why design knobs such as interlayer thickness, resistivity–temperature parameters, and grid resolution directly shape minimum sensitivity, OMP recovery, and NET.

This viewpoint also suggests a connection to asynchronous temporal-contrast sensors such as the device of Lichtsteiner \emph{et al.}~\cite{lichtsteiner2008128}. In that work, each pixel generates address-events when the relative optical signal changes, i.e., the contrast signal is proportional to \(d\ln I_{\mathrm{ph}}/dt\), where \(I_{\mathrm{ph}}\) denotes photocurrent, and the event output is realized by digital circuitry and an address-event representation (AER) interface. In ThermoMesh, an analogous event-driven interpretation could emerge from the temperature-dependent nonlinear interlayer response if one considers relative boundary-voltage changes rather than relative photocurrent changes: the relevant contrast-like quantity would be based on \(d\ln V/dt\) at the boundary channels, while the thresholding mechanism would be realized primarily by the nonlinear material response rather than by in-pixel digital circuits. This analogy suggests a plausible path toward future event-driven ThermoMesh readout.

Building on this hardware-as-computation viewpoint, the way ThermoMesh multiplexes information naturally connects to compressive sensing ideas. 
This work is conceptually similar to snapshot compressive imaging~\cite{yuan2021snapshot}, but instead of compressing higher-dimensional information into 2D, it compresses 2D information into 1D. Classic snapshot systems multiplex a spatial field into a lower-dimensional measurement in a single shot; here, the resistive/thermoelectric network multiplexes the interior temperature field to a perimeter vector of boundary voltages. The sparsity prior on \( \mathbf{T} \) supplies the missing constraints for inversion, which explains the role of the NSP (linear case), OMP-based recovery, and the observed improvements in classification accuracy and MAE once the interlayer enhances separability at the boundary.

To move from conceptual analogy to concrete design choices, we will require a simulation workflow that captures both physics and computation. 
Simulating the computation carried out by a nonlinear resistive network~\cite{scellier2024fast,scellier2025universal} provides a principled way to explore interlayer physics and geometry before fabrication. Our KCL-based assembly of node equations already yields \( \mathbf{A} \) and \( \mathbf{A}(\mathbf{T}) \); extending this with established nonlinear network solvers enables fast, convergent updates of node potentials and currents as material laws \( \rho(T) \) and geometry parameters vary. This workflow makes it possible to sweep interlayer thickness, \(\beta\) in the NTC law, and grid size to quantify how these choices reshape minimum sensitivity, uniformity, and NET across operating regimes.

With an efficient simulator in hand, the learning rule should align with the device’s native relaxation dynamics. Training the network as an energy-based model using Equilibrium Propagation~\cite{kendall2006training} aligns naturally with the device physics. The boundary electrodes play the role of output nodes: in the free phase they are measured while left electrically floating, and in the nudged phase they can be weakly perturbed by injected currents, while interior node voltages relax to steady states that minimize an energy functional defined by conductances and Seebeck sources. Equilibrium Propagation leverages this physical relaxation both for inference and for learning local parameters, suggesting a co-design loop in which material parameters (e.g., \( \rho_0 \), \( \beta \)) and readout weights are tuned using gradient information derived from the same dynamics that govern operation. We envision the realization of this concept through hardware-in-the-loop training at the manufacturing stage.

These modeling and learning ingredients, taken together, point to a spectrum of sensing architectures that balance deterministic and random structures. In the future, we seek to discover a context-aware structured nonlinear sensing matrix~\cite{akhtar2015efficient} that lies between deterministic cross-bar array~\cite{aguirre2024hardware} designs and purely random reservoirs~\cite{zhu2023online,qi2023physical}. The results in Section~\ref{sec:Design and Performance} already trace this continuum: a regular cross-bar supports analytic guarantees (e.g., NSP for \(q{=}1\)), whereas measured nonlinear interlayers introduce controllable nonlinearity that improves separability under noise. Introducing mild structure in the nonlinearity (patterned thickness, graded \(\beta\), or sparse cross-links) promises reservoir-like richness while preserving uniqueness guarantees and the low-power readout advantages of a deterministic array.

The design of the linear sensing matrix could, in principle, be formulated as a modified overcomplete dictionary-learning problem~\cite{patel2011sparse}—where, instead of enforcing sparsity on the input signal, we assume known input sparsity and impose structural constraints on the sensing matrix (i.e., on the sensor design).

The read-only resistive memory programmed at the manufacturing stage is conceptually identical to the Mask ROM~\cite{leblebici1996cmos} used in the early stages of digital computers. This decision is justifiable for context-specific applications and also improves the reliability of the sensor. In our setting, “programming” consists of choosing materials, thicknesses, and layouts to set the entries of \( \mathbf{A} \) once and for all. For embedded, long-lifetime deployments—where calibration drift and write-induced variability are unacceptable—this Mask-ROM-like choice trades reconfigurability for predictability, environmental robustness, and consistent sensitivity and NET over time.

\begin{figure}[!t]
    \centering

    \includegraphics[width=\linewidth,height=0.36\textheight,keepaspectratio]{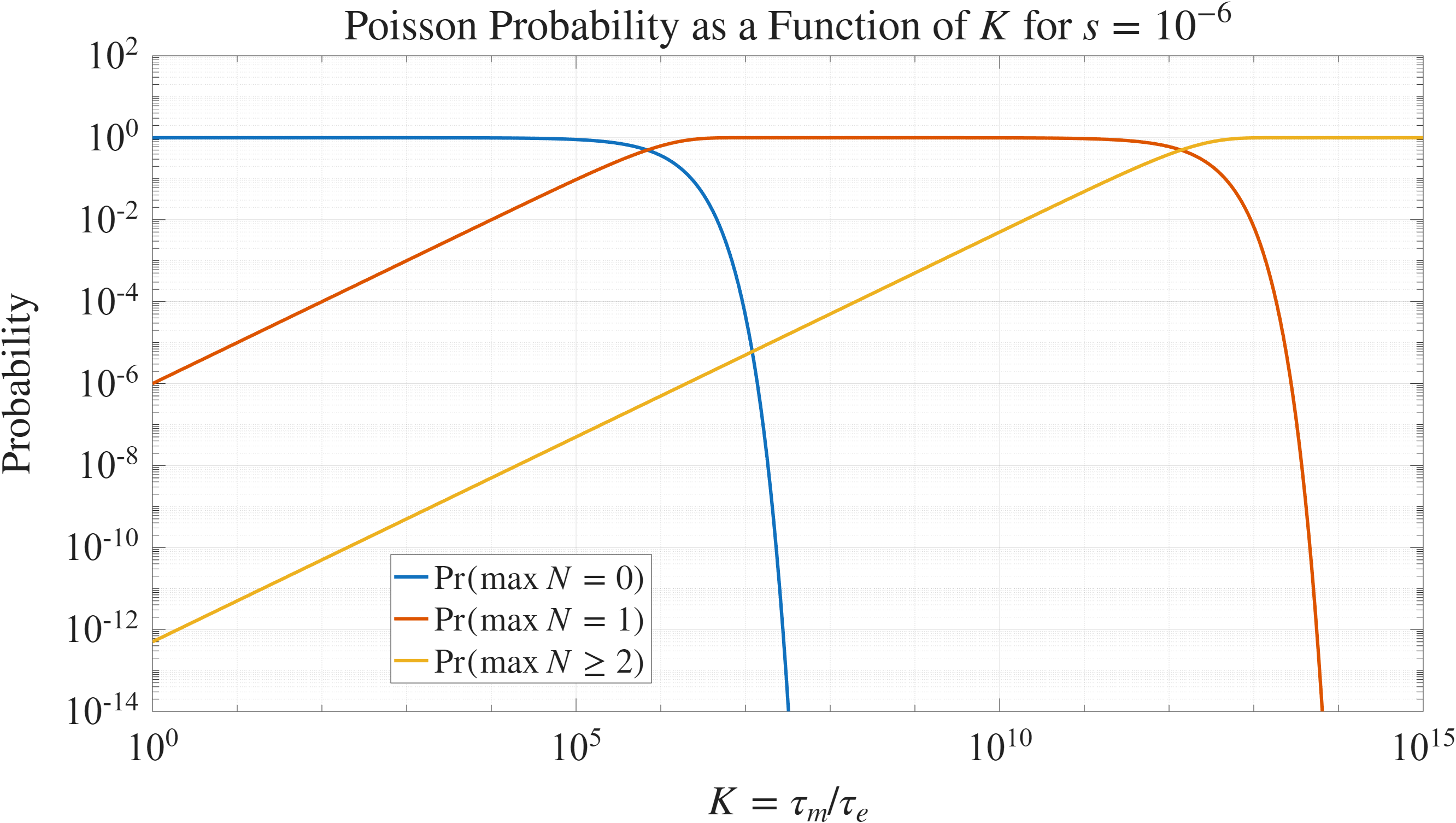}
    \caption*{(a)}

    \vspace{5mm}

    \includegraphics[width=\linewidth,height=0.36\textheight,keepaspectratio]{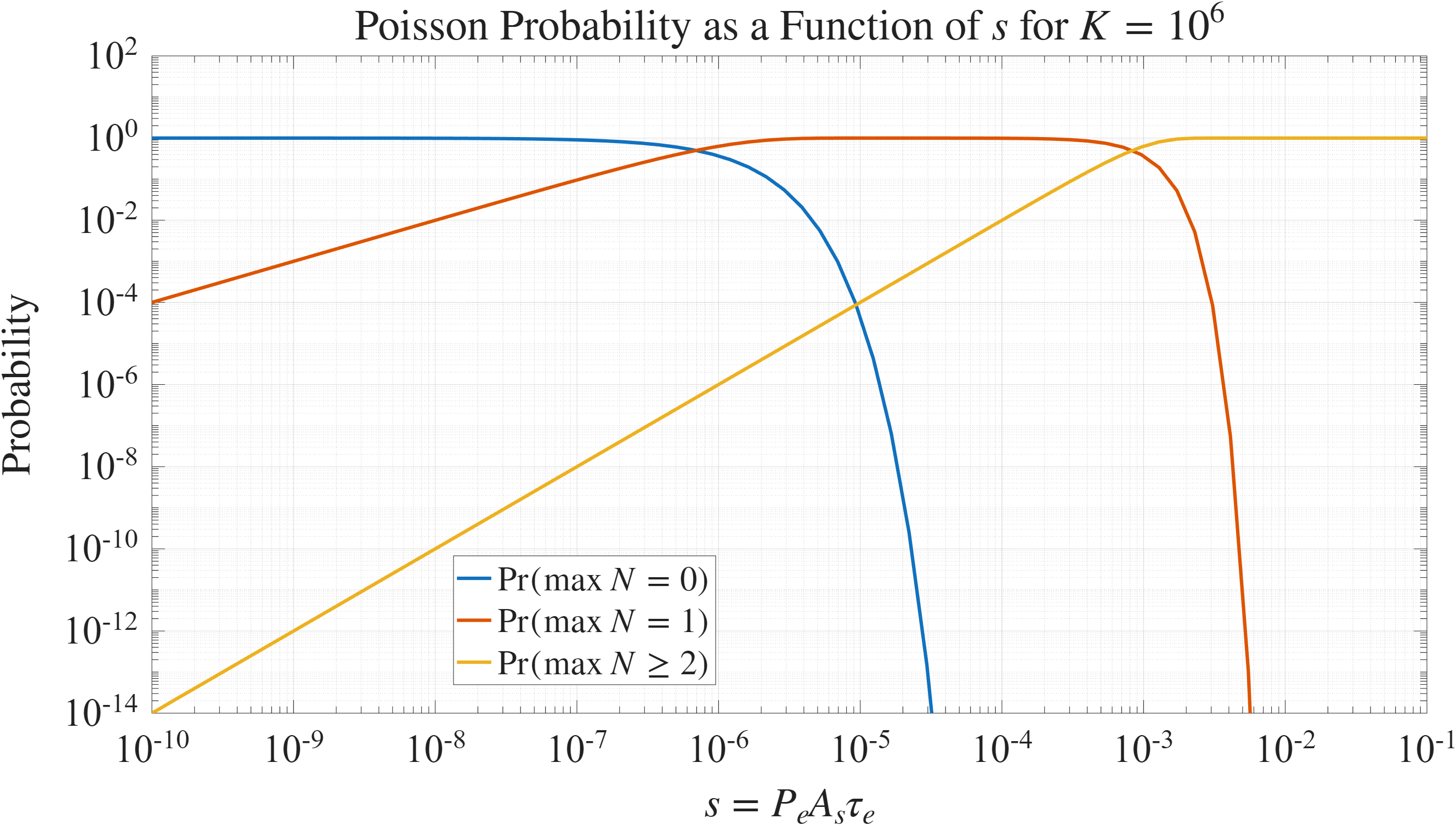}
    \caption*{(b)}

    \caption{\textbf{Windowed Poisson overlap model.}
    Probabilities that the \emph{maximum} number of simultaneous events within a measurement window is $0$, $1$, or $\ge 2$.
    (a) Dependence on the window ratio $K=\tau_m/\tau_e$ for a fixed Poisson mean $s$.
    (b) Dependence on the Poisson mean $s$ for a fixed window ratio $K$.
    These curves quantify the likelihood of sparsity violation due to event overlap.}
    \label{fig:poisson-windowed}
\end{figure}

After defining temporal sparsity deterministically in Section~\ref{sec:design-metrics}, we evaluate here whether a given sensor design is likely to satisfy a characterized temporal sparsity range under random event arrivals. Let $N_{\max}$ denote the maximum number of simultaneous events present within a measurement window. For a sensor characterized up to temporal sparsity $q_t^{\max}$, an operating condition is regarded as acceptable at tolerance $\delta$ if
\begin{equation}
  \Pr\!\left\{N_{\max} \ge q_t^{\max}+1\right\} \le \delta,
\end{equation}
where $\delta$ is a user-specified overlap/failure tolerance. The key implication here is that the likelihood of violating the characterized temporal sparsity range increases with the window ratio, $K \approx \tau_m/\tau_e$, so longer measurement windows are more likely to contain overlapping events. This behavior is illustrated in Fig.~\ref{fig:poisson-windowed} using the Poisson overlap model described in~\ref{app:poisson}.

For Eq.~(25),~\ref{app:poisson} provides the explicit propagation from application statistics to temporal-sparsity tolerance. In particular, Eq.~\eqref{eq:sparsity-scaling} gives $s=P_eA_s\tau_e$, and Eq.~\eqref{eq:windowed} then maps $s$ and $K$ to the overlap probability. Since $P_e$ is application dependent, it is more informative here to invert that calculation and report the maximum areal event rate compatible with a chosen tolerance. For a contact-mode $16\times16$ ThermoMesh with $50~\mu\mathrm{m}$ pitch, the sensing area is approximately $A_s=(16\times50~\mu\mathrm{m})^2=6.4\times10^{-7}~\mathrm{m}^2$. Taking $q_t^{\max}=1$, $\tau_e\approx1~\mathrm{ms}$, and $K=10^{6}$, Eq.~\eqref{eq:windowed} gives $\Pr\{N_{\max}\ge2\}\le1\%$ provided $P_e\lesssim2.2\times10^{5}~\mathrm{m}^{-2}\mathrm{s}^{-1}$. Tightening the tolerance to $10^{-4}$ reduces this bound to $P_e\lesssim2.2\times10^{4}~\mathrm{m}^{-2}\mathrm{s}^{-1}$. This makes explicit how an assumed rare-event arrival rate is converted into a confidence that 1-sparse operation is satisfied.

\begin{figure}[!t]
  \centering
  \includegraphics[width=\linewidth,height=\textheight,keepaspectratio]{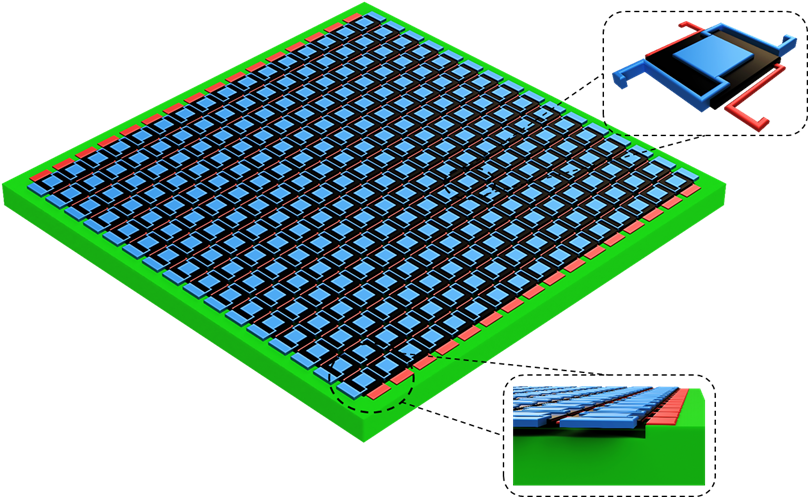}
  \caption{\textbf{Conceptual} bolometer-style isolated-pixel variant used to support effectively 1-sparse frames.}
  \label{fig:bolometer-schematic}
\end{figure}

Fig.~\ref{fig:bolometer-schematic} illustrates a bolometer-style realization of the sparse-event assumptions. In that non-contact mode, it is useful to view the pixel geometry through a simple fill factor,
\begin{equation}
\text{fill factor}=\frac{A_p}{A_s/(MN)},
\label{eq:fillfactor}
\end{equation}
which should be large enough that the event footprint satisfies $A_e \ll A_p$. Thermal isolation then suppresses lateral conduction and, with $\tau_s \ll \tau_e$, each event can be treated as quasi-static at the pixel level. For a $16\times16$ bolometer-style array with $A_p=(10~\mu\mathrm{m})^2$ and a fill factor $=0.9$, the sensing area is $A_s=(MN)A_p/0.9\approx2.84\times10^{-8}~\mathrm{m}^2$. Using the same illustrative values $q_t^{\max}=1$, $\tau_e\approx1~\mathrm{ms}$, and $K=10^{6}$, Eq.~\eqref{eq:windowed} gives $P_e\lesssim5.0\times10^{6}~\mathrm{m}^{-2}\mathrm{s}^{-1}$ for $\delta=1\%$ and $P_e\lesssim5.0\times10^{5}~\mathrm{m}^{-2}\mathrm{s}^{-1}$ for $\delta=10^{-4}$. The smaller sensing area therefore relaxes the temporal-overlap constraint at a fixed areal event rate. Separately, because $A_e \ll A_p$ in this bolometer limit, the fill factor also controls capture probability: for a uniformly distributed subpixel event, a $90\%$ fill factor corresponds to an approximately $90\%$ probability that the event lands on the active absorber within a unit cell. By contrast, in the contact-mode ThermoMesh emphasized in this work, the relevant 1-sparse condition is that the detectable heated footprint satisfy $A_e \gg A_p$ while still exciting only one pixel. Thus, the two modes satisfy the same one-pixel-per-frame requirement through different geometries and heat-spreading physics.

As a concrete use case, the finalized \(16\times16\) ThermoMesh with \(50~\mu\mathrm{m}\) junction pitch is well matched to high-temperature molten-droplet detection. For this contact-mode geometry, Eq.~\eqref{eq:sparsity-scaling} directly maps the assumed droplet-arrival rate to the temporal-sparsity tolerance in Eq.~\eqref{eq:windowed}; the illustrative bounds above show how a specified overlap tolerance can be converted into an allowable operating regime. Under the ceramic-NTC interlayer, the sensitivity analysis showed a substantial improvement in the minimum sensitivity under elevated-temperature conditions relative to the uniform ambient-temperature case (see Section~\ref{sec:Design and Performance}). Together with the boundary-only readout and the demonstrated localization, regression, and NET performance (Tables~\ref{tab:ml-acc-net}--\ref{tab:ml-mae}), these results indicate that the reported design is practically useful for detecting and characterizing small, fast molten droplets in embedded or harsh environments where optical access is limited.

\section{Conclusions and Future Work}\label{sec:Conclusion}

This paper introduced \emph{ThermoMesh}, a passive digital–-analog thermoelectric thin-film sensor that maps a spatio–temporally sparse temperature field \(\mathbf{T}\) to boundary voltages \(\mathbf{V}\) through Seebeck transduction and a resistive interlayer. Focusing on the single-event (1-sparse) regime, we defined four performance metrics---range, efficiency, sensitivity, and accuracy---and used them to compare baseline, linear-resistive, and nonlinear-interlayer designs. In the linear regime, a constant-resistance interlayer increased the minimum sensitivity and improved spatial uniformity. By comparison, an NTC interlayer operated as a non-ideal thermal switch and sustained higher minimum sensitivity as resolution grew, thereby preserving distinguishability at scale. The temperature-dependent nonlinear interlayer response also suggests a possible future extension toward event-driven boundary readout based on relative voltage transients. Using synthetic data generated by the measurement models, the learning pipeline achieved high localization accuracy, low MAE for temperature regression, and reduced NET. As one practical application, the reported design (e.g., a \(16\times16\) mesh with \(50~\mu\mathrm{m}\) pitch and a ceramic-NTC interlayer) appears well suited to high-temperature molten-droplet detection in settings where optical access is limited. In this context, the observed \emph{misclassifications } are typically adjacent to the true hot pixel, which is sufficient to trigger event capture, guide a localized inverse heat-conduction refinement around the impact site, and reliably estimate temperature at the correct region of interest.

Looking ahead, we will experimentally validate ThermoMesh and rigorously quantify sensor resolution, including point-spread/transfer functions and the minimum resolvable separation between two sources as a function of SNR and interlayer design, while incorporating realistic noise and parasitic interlayer capacitance into the NET analysis and end-to-end transient simulations to quantify any RC-limited effect on response time. We will exploit the temporal structure of the perimeter stream by coupling quickest change-point detection \cite{xie2012change} for statistically grounded event triggers with inverse heat-conduction solvers for rapid physics-based refinement, enabling event-driven operation in embedded environments. On the modeling side, we will extend the nonlinear setting by coupling physics-informed machine learning \cite{karniadakis2021physics} with emerging nonlinear compressive sensing formulations \cite{blumensath2013compressed}. Guided by inverse modeling~\cite{liu2024data}, we will expand the materials search space beyond ceramic NTCs and VO\(_2\) by incorporating high-thermopower transverse thermoelectrics, pyroelectric crystals with larger \(RC\) time constants to separate fast and slow dynamics in a material-programmed manner, analogous in spirit to recent architectural physics embeddings for multiscale dynamics~\cite{kohli1997pyroelectric,zhang2026physics}, molecular thermal switches with extreme on/off ratios, and flexible neuromorphic nanomaterials as learnable priors in the training loop~\cite{daus2022fast,feng2024transverse,li2023electrically,ding2024nanomaterials}, thereby co-designing geometry, materials, dynamics, and readout for application-specific operating ranges under fabrication constraints.

\appendix
\section{Rare-event model and window validity}\label{app:poisson}

To characterize random events, we model events as a homogeneous Poisson process with average areal rate $P_e$ (units: m$^{-2}$s$^{-1}$). The expected number of \emph{simultaneous} ongoing events over the whole sensor at a random instant is
\begin{equation}
  s \;=\; P_e\,A_s\,\tau_e ,
  \label{eq:sparsity-scaling}
\end{equation}
so $s$ is the mean number of events present in a single event-length snapshot over area $A_s$. In a single snapshot of duration $\tau_e$, a homogeneous Poisson model gives
\begin{equation}
  \Pr\{N=n\} \;=\; e^{-s}\,\frac{s^{n}}{n!},
  \qquad n=0,1,2,\ldots,
  \label{eq:poi-exact}
\end{equation}
regardless of how we choose to sample or aggregate in time. For $s\ll 1$ we have
\begin{equation}
  \Pr\{N=1\}\approx s, \qquad
  \Pr\{N\ge 2\}\approx \frac{s^2}{2}\ll s,
  \label{eq:Pr =s}
\end{equation}
so multi-event overlap is intrinsically rare when $s$ is small.

ThermoMesh readout corresponds to a \emph{measurement window} of length $\tau_m$. For a given event duration $\tau_e$, the ratio
\begin{equation}
  K \;\approx\; \frac{\tau_m}{\tau_e}
\end{equation}
is the number of event-length segments that fit inside one measurement window. We consider the random process $N(t)$, the number of simultaneous active events, and define the windowed maximum
\begin{equation}
  N_{\max} \;=\; \max_{t\in[0,\tau_m]} N(t).
  \label{eq:N_Max}
\end{equation}

For a sensor characterized up to temporal sparsity $q_t^{\max}$, a window is considered \emph{valid} if $N_{\max}\le q_t^{\max}$ and \emph{invalid} if $N_{\max}\ge q_t^{\max}+1$. Approximating the window as $K$ statistically independent snapshots of duration $\tau_e$ leads to
\begin{equation}
  \Pr\!\big\{N_{\max}\ge q_t^{\max}+1\big\}
  \;\approx\;
  1-\!\left(\sum_{n=0}^{q_t^{\max}} e^{-s}\frac{s^n}{n!}\right)^{K},
  \label{eq:windowed}
\end{equation}
which is the invalidation (failure) probability for a sensor characterized up to temporal sparsity $q_t^{\max}$.

For the $q_t^{\max}=1$ case we define
\[
  P_0(K,s) = \Pr\{N_{\max} = 0\},\quad
  P_1(K,s) = \Pr\{N_{\max} = 1\},\quad
  P_{\ge2}(K,s) = \Pr\{N_{\max} \ge 2\},
\]
so that $P_{\ge2}$ is the failure probability for a sensor characterized up to temporal sparsity $1$, and $P_1$ measures the fraction of windows that are both non-empty and valid for that case.

\section*{Acknowledgements}
We thank Scott Schiffres, Dehao Liu, Wenfeng Zhao, and Bruce Murray for valuable discussions and constructive feedback on this work.

\section*{Nomenclature}
\begin{longtable}{@{}p{0.20\linewidth}p{0.76\linewidth}@{}}

$\mathbf{A}$ &
Sensitivity matrix mapping junction-temperature vector $\mathbf{T}$ to boundary-voltage vector $\mathbf{V}$ \\

$A_e$ &
Event footprint area \\

$A_p$ &
Pixel (junction) transducing area \\

$A_s$ &
Total sensor (sensing) area \\

$d_{\rm norm}$ &
Normalized Euclidean localization error \\

$E_s$ &
Readout energy required to acquire one boundary-channel voltage sample \\

$\mathbf{G}$ &
Inter-node conductance matrix in the KCL network model \\

$I_{\mathrm{ph}}$ &
Photocurrent used in the temporal-contrast sensor analogy \\

$\mathbf{I}_U,\mathbf{I}_S,\mathbf{I}_V$ &
Selection/reference matrices used in boundary-voltage computation \\

$K$ &
Window ratio ($K\approx\tau_m/\tau_e$) \\

$M$ &
Number of junction (pixel) rows in the $M\times N$ ThermoMesh grid \\

$N$ &
Number of junction (pixel) columns in the $M\times N$ ThermoMesh grid \\

$P_e$ &
Areal event rate (events\,m$^{-2}$\,s$^{-1}$) \\

$P_n$ &
Boundary-noise power used in the SNR-based noise model \\

$P_s$ &
Boundary-signal power used in the SNR-based noise model \\

$q_s^{\max}$ &
Maximum spatial sparsity level up to which the sensor is characterized \\

$q_t^{\max}$ &
Maximum temporal sparsity level up to which the sensor is characterized \\

$R$ &
Interlayer resistance \\

$s$ &
Poisson mean number of simultaneous ongoing events over $A_s$ in an event-length snapshot ($s=P_eA_s\tau_e$) \\

$S$ &
Effective Seebeck coupling coefficient(s) in the network model \\

$t_n$ &
Sampling time associated with frame index $n$ \\

$T$ &
Junction temperature on the $M\times N$ grid \\

$\mathbf{U}$ &
Interior node electric potentials \\

$\mathbf{V}$ &
Measured boundary-voltage vector \\

$\beta$ &
Thermistor $\beta$ (B-parameter / beta constant) used in the exponential NTC model \\

$\delta$ &
Allowed overlap/failure probability used to evaluate whether random operating conditions satisfy the characterized temporal sparsity range \\

$\varepsilon_0$ &
Permittivity of free space \\

$\varepsilon_r$ &
Relative permittivity of the resistive interlayer \\

$\eta_{\mathrm{ch}}$ &
Channel-count reduction factor \\

$\kappa$ &
Super-linearity factor of boundary response \\

$\boldsymbol{\nu}$ &
Boundary-voltage noise vector \\

$\rho$ &
Interlayer resistivity \\

$\sigma$ &
Sensitivity (boundary-voltage swing per kelvin) \\

$\tau_e$ &
Event duration \\

$\tau_m$ &
Measurement-window duration \\

$\tau_s$ &
Pixel characteristic thermal equilibration time constant \\

\end{longtable}

\section*{Abbreviations}
\begin{longtable}{@{}p{0.18\linewidth}p{0.78\linewidth}@{}}
AER  & Address--event representation \\

EMF  & Electromotive force \\

KCL  & Kirchhoff's current law \\

LSTM & Long short-term memory \\

MAE  & Mean absolute error \\

MIT  & Metal--insulator transition \\

MSE  & Mean squared error \\

NET  & Noise-equivalent temperature \\

NSP  & Null space property \\

NTC  & Negative temperature coefficient \\

OMP  & Orthogonal matching pursuit \\

SNR  & Signal-to-noise ratio \\
\end{longtable}

\section*{Funding sources}
This research was supported by the Seed Grant Program of the Smart Energy Transdisciplinary Area of Excellence (SE TAE) at Binghamton University.

\section*{Declaration of Competing Interest}
The authors declare that they have no known competing financial interests or personal relationships that could have appeared to influence the work reported in this paper.

\bibliography{mybibfile}

\end{document}